\pdfoutput=1

\documentclass[11pt]{article}

\usepackage{math}
\usepackage[preprint]{acl}

\usepackage{times}
\usepackage{latexsym}

\usepackage[T1]{fontenc}

\usepackage[utf8]{inputenc}

\usepackage{microtype}

\usepackage{inconsolata}

\title{On Eliciting Syntax from Language Models via Hashing}

\author{Yiran Wang\quad Masao Utiyama \\
        National Institute of Information and Communications Technology (NICT) \\
        \texttt{yiran.wang@nict.go.jp\quad mutiyama@nict.go.jp}}

\begin{document}
\maketitle

\begin{abstract}
Unsupervised parsing, also known as grammar induction, aims to infer syntactic structure from raw text. Recently, binary representation has exhibited remarkable information-preserving capabilities at both lexicon and syntax levels. In this paper, we explore the possibility of leveraging this capability to deduce parsing trees from raw text, relying solely on the implicitly induced grammars within models. To achieve this, we upgrade the bit-level CKY from zero-order to first-order to encode the lexicon and syntax in a unified binary representation space, switch training from supervised to unsupervised under the contrastive hashing framework, and introduce a novel loss function to impose stronger yet balanced alignment signals. Our model\footnote{\url{https://github.com/speedcell4/parserker}} shows competitive performance on various datasets, therefore, we claim that our method is effective and efficient enough to acquire high-quality parsing trees from pre-trained language models at a low cost.
\end{abstract}

\section{Introduction}

\begin{figure}[t]
    \centering
    \includegraphics[width=\columnwidth]{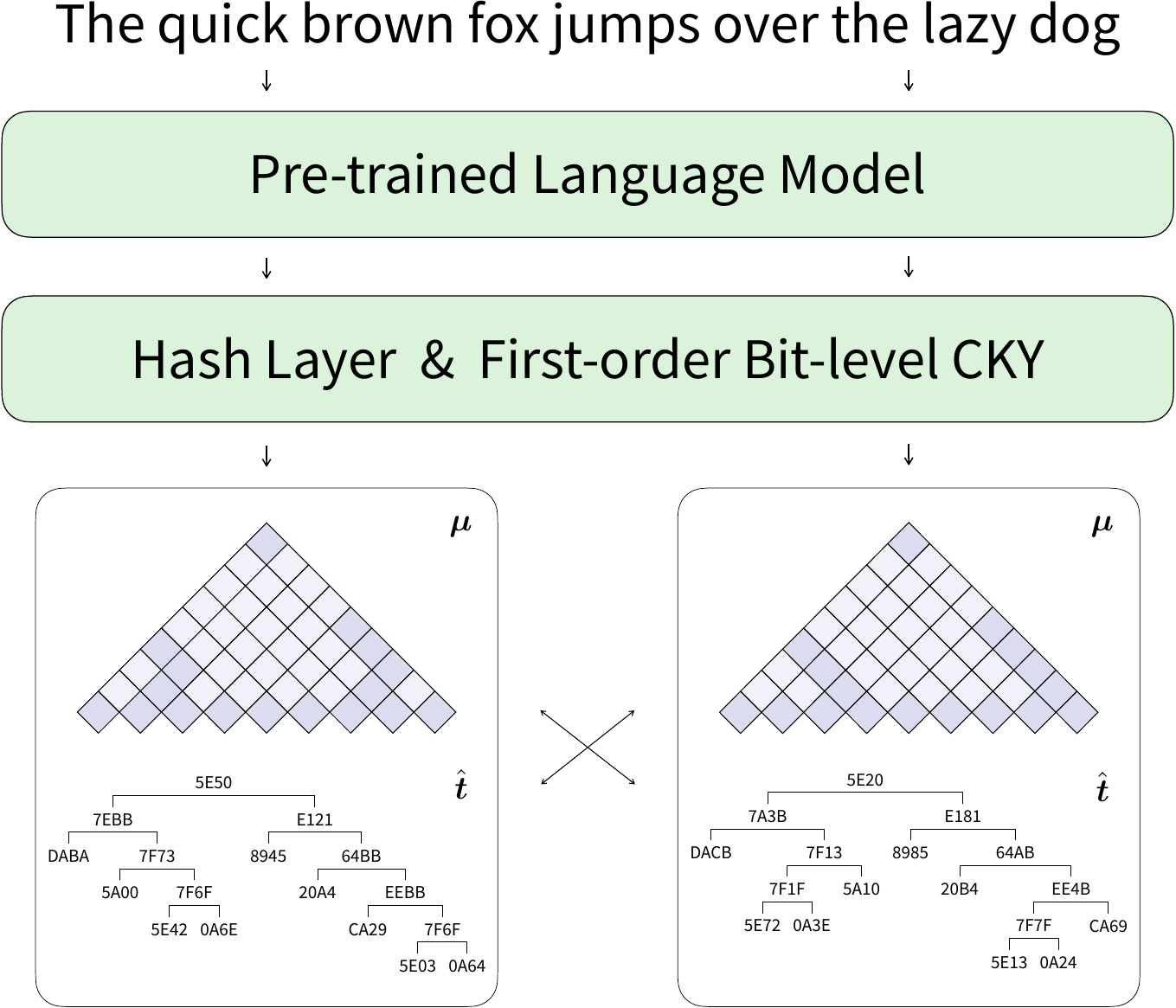}
    \caption{The model architecture. The hash layer produces scores of all spans, and the following first-order bit-level CKY (\S\ref{sec:first_cky}) returns marginal probabilities $\ve{\mu}$ and predicts the most probable trees $\hat{\ve{t}}$. Sentences are fed into the network twice, We select span marginal probabilities from one pass according to the predicted trees from the other pass, and perform contrastive hashing (\S\ref{sec:unsup_hash}, \S\ref{sec:loss}) on their corresponding score and code vectors. The purple cells represent the marginal probabilities, and the dark purple indicate the selected ones.}
    \label{fig:arch}
\end{figure}

Grammars form the backbone of languages, providing the essential framework that dictates how lexicons are arranged to convey meaning. Understanding and generating language heavily relies on grasping these latent structures. Unsupervised parsing, which aims to deduce sentence structure without relying on costly manually annotated treebanks, has been widely studied in academia. However, despite its importance, advancements have been slow due to the intrinsic complexity of this task. Nowadays, addressing these challenges becomes even more crucial for further exploring the capabilities of large language models.

Word embedding and language model techniques \cite{NIPS2013_9aa42b31,mikolov-etal-2013-linguistic,radford2018improving,devlin-etal-2019-bert} have shown that training models to predict tokens in specific contexts is remarkably effective in implicitly capturing lexical features. A well-known example is the captured lexical relationship of \emph{king - man + woman = queen}. As one of the most widely accepted explanations for this phenomenon, the distributional hypothesis \cite{doi:10.1080/00437956.1954.11659520,NIPS2013_9aa42b31,mikolov-etal-2013-linguistic} suggests this is because tokens appearing in similar contexts tend to be assigned similar meanings. Specifically, similar contexts achieve this by placing tokens in analogous syntactic structures. This phenomenon naturally prompts us to consider whether there is a representation learning method that can explicitly encode both lexical and syntactic information in a unified format, making it possible to capture syntactic structures as well as lexical relationships by training language models solely with conventional conditional token prediction procedures.

Fortunately, the recently proposed binary representation meets these requirements perfectly. \citet{wang-etal-2023-24} proposed a binary representation that bridges the gap between the continuous nature of deep learning and the discrete intrinsic property of natural languages. Instead of directly applying contrastive learning on the high-dimensional continuous hidden states of pre-trained language models, \citet{wang-etal-2023-24} project them as $K$-dimensional score vectors. These scores can easily be binarized into $K$-bit codes, and token-level contrastive learning is applied among these scores and their binarized codes. They demonstrate that lexical information can be properly preserved within only 24 bits. Following this, \citet{wang-utiyama-cont-disc} additionally take spans on the target parsing trees into consideration. They use marginal probabilities to construct a novel similarity function that reflects not only lexical information but also the boundary of each span, and then perform contrastive hashing across spans rather than tokens. In their supervised parsing experiments, they show the effectiveness of the structured binary representation by achieving comparable performance to conventional parsers.

Although in the supervised settings, \citet{wang-utiyama-cont-disc} achieves satisfactory results, we found that for unsupervised settings, their model is insufficient to induce meaningful parsing trees. In this paper, we aim to elicit constituency parsers from pre-trained language models without training them on annotated treebanks. We analyze the existing issues of their structured binary representation and explore the possibility of further enhancing the unified information-preserving capability. To achieve this, we upgrade the bit-level CKY module from zero-order (\S\ref{sec:zero_cky}) to first-order (\S\ref{sec:first_cky}) to integrate lexicon and syntax in a unified format, convert parsing from supervised (\S\ref{sec:sup_hash}) to unsupervised (\S\ref{sec:unsup_hash}), and propose a novel objective function (\S\ref{sec:loss}) to impose stronger yet balanced alignment signals. Besides, we also discuss how the learning objective of contrastive hashing aligns with the target of parsing. This provides an explanation (\S\ref{sec:unsup_hash}) different from the distributional hypothesis and explains why our training leads to syntactic structures rather than other structures. Experiments show that our models achieve competitive performance and indicate that acquiring high-quality syntactic annotations at a low cost is becoming practicable. We refer to our parser as \texttt{Parserker2}, following the original \texttt{Parserker} \cite{wang-utiyama-cont-disc}.

\section{Background}
\label{sec:background}

\subsection{Zero-order Constituency Parsing}
\label{sec:zero_cky}

Given sequence $w_1,\dots,w_n$, constituency parser returns the most probable binary-branching parsing tree $\ve{t} = \set{\agl{l_i,r_i,y_i}}_{i=1}^{2n-1}$, which is represented as a list of labeled spans indicating constituents at different hierarchies. Where $l_i$ and $r_i$ refer to the left and right boundaries of the $i$-th constituent, and $y_i\in\ca{y}$ stands for its assigned label. Previous models \cite{kitaev-klein-2018-constituency,yu-etal-2020-named,ijcai2020p0560} commonly employ encoders to transform inputs into hidden states $\ve{h}_1,\dots,\ve{h}_n$ first, use classifiers to predict span scores $g(l,r,y)$ and tree scores $g(\ve{t})$, and then normalize them among all valid trees to obtain tree probability $p(\ve{t})$. Training and inference stages aim at maximizing the probabilities of target trees and searching trees with the maximal probabilities $\prd{\ve{t}}$, respectively. \begin{gather}
    g(\ve{t}) = \sum_{\agl{l,r,y}\in\ve{t}}{g(l,r,y)} \vh \\
    p(\ve{t}) = \dfrac{\exp{g(\ve{t})}}{Z\equiv\sum_{\ve{t}'\in\ca{t}}{\exp{g(\ve{t}')}}} \label{eq:prob} \vh \\
    \prd{\ve{t}} = \set{\agl{l_i,r_i,y_i}}_{i=1}^{2n-1} \gets \argmax_{\ve{t}\in\ca{t}}{~p(\ve{t})} \vh 
\end{gather} 

Apart from being used to normalize probabilities of trees, the log partition term $Z$, which stands for the total scores of all valid constituency trees, can also be used to compute span marginal probabilities. As \citet{eisner-2016-inside} mentioned, computing the partial derivative of the log partition with respect to span scores yields marginal probabilities efficiently. \begin{equation}
        \mu{(l,r,y)} = \dfrac{\partial\,\log{Z}}{\partial\,g(l,r,y)} 
\end{equation}

Intuitively speaking, marginal probability reflects the joint probability of selecting tokens $w_l, \dots, w_r$ as a constituent with label $y$ assigned to it. If a span is not likely to be selected, its marginal probability will not be high regardless of its label. Therefore, similar to hidden states, marginal probabilities are considered a format containing not only lexical but also syntactic features. Unlike hidden states, these marginal probabilities explicitly correspond to the specific boundaries and labels of spans in parsing trees globally normalized under the CKY framework, whereas hidden states implicitly preserve this information in a high-dimensional, human-unreadable format.

Recently, \citet{wang-utiyama-cont-disc} extended constituency parsers by replacing discrete labels $y \in \ca{y}$ with binary codes $\ve{c} \in \B^{K}$. In their approach, the code-level scores $g(l, r, \ve{c})$ are obtained by summing up bit-level scores $g_k(l, r, c^k)$. \begin{gather}
    g(\ve{t}) = \sum_{\agl{l,r,\ve{c}}\in\ve{t}}{g(l,r,\ve{c})} \\
    g(l,r,\ve{c}) = \sum_{k=1}^K{g_k(l,r,c^k)} 
\end{gather}

Moreover, to compute these bit-level scores, they retained the one-head-one-bit design of \citet{wang-etal-2023-24} and employed a multi-head attention module to predict the score of being assigned $+1$. \begin{align}
    g_k(l,r,+1) &= \dfrac{(\ma{w}_k^Q\ve{h}_{l})^{\top}(\ma{w}_k^K\ve{h}_{r})}{\sqrt{d_k}} \label{eq:zero_g} \\
    g_k(l,r,-1) &= 0
\end{align} Where $\ma{w}_k^Q, \ma{w}_k^K \in \R^{\lceil \frac{d}{K} \rceil \times d}$ are the query and key matrices used to produce the $k$-th bit. They assign a score of $0$ for the $-1$ case and extend the marginal probability and decoding to the bit level. \begin{gather}
    \mu_k{(l,r,c^k)} = \dfrac{\partial\,\log{Z}}{\partial\,g_k(l,r,c^k)} \vh \\
    \prd{\ve{t}} = \set{\agl{l_i,r_i,\ve{c}_i}}_{i=1}^{2n-1} \gets \argmax_{\ve{t}\in\ca{t}}{~p(\ve{t})} \vh
\end{gather}

\subsection{Supervised Contrastive Hashing}
\label{sec:sup_hash}

To perform contrastive learning, \citet{wang-etal-2023-24} and \citet{wang-utiyama-cont-disc} define their similarity functions in a similar manner, both first binarize one input and then calculate the similarity between the continuous one and the binarized one. However, the former binarizes scores via taking their signs, while the latter leans bits towards the sides with larger marginal probabilities. \begin{gather}
    \ve{c} = [c^1,\dots,c^K] \in \B^{K} \\
    c^k = \begin{cases}
        +1 & \mu_k{(l,r,+1)} > \mu_k{(l,r,-1)} \nonumber \\
        -1 & \text{otherwise}
    \end{cases}
\end{gather}

As mentioned above, marginal probabilities contain both label and structural information. To impose supervision on lexicon and syntax simultaneously by leveraging this property, they proposed defining the novel similarity as the average of bit-level marginal probabilities of the $i$-th constituent with the binary label of $j$-th constituent. \begin{gather}
    s(i,j) = \dfrac{1}{K}\sum_{k=1}^K{~\mu_k(l_i,r_i,c_j^k)}
\end{gather}

During the training stage, \citet{wang-utiyama-cont-disc} select spans from target trees to perform contrastive hashing with the similarity function described above. Naively contrasting all spans would increase the time complexity to $\mathcal{O}(n^4)$. To avoid this, they restrict supervision to spans on the target trees, reducing the number of spans to $2n-1$ and maintaining the time complexity at $\mathcal{O}(n^2)$. In their supervised settings, they only allow the model to determine the binary codes, without predicting the boundaries, thus, the procedure can be reinterpreted as searching in a constrained space. \begin{gather}
    \prd{\ve{t}}=\set{\agl{l_i,r_i,\prd{\ve{c}}_i}}_{i=1}^{2n-1} \gets \argmax_{\ve{t}\in\ca{t}{[\ve{l},\ve{r},\cdot]}}{~p(\ve{t})} \label{eq:sup_argmax}
\end{gather} Where $l_i$ and $r_i$ denote the boundaries of the target spans, and $\ca{t}{[\ve{l},\ve{r},\cdot]}$ means only searching in the constrained space to ensure target are always included. Besides, the positive and negative sets are divided according to the ground-truth labels $y_i$. \begin{equation}
\begin{gathered}
    \ca{p}=\set{j\mid y_i= y_j} \\
    \ca{n}=\set{j\mid y_i\not= y_j}
\end{gathered}
\label{eq:sup_np}
\end{equation}

\section{Proposed Methods}
\label{sec:proposed}

\begin{figure}[t]
    \centering
    \includegraphics[width=\columnwidth]{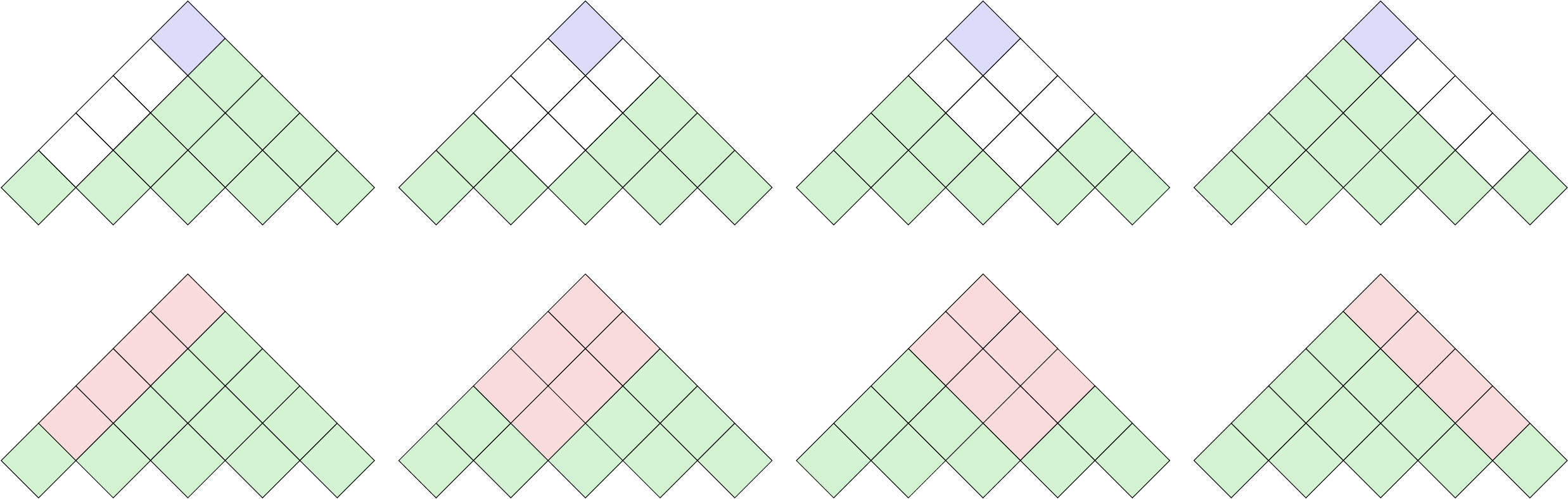}
    \caption{Charts of the zero-order (above \S\ref{sec:zero_cky}) and the first-order parsing (below \S\ref{sec:first_cky}). At this time step, zero-order parsers separately determine the splitting positions on the \textgreen{left and right children} and predict labels according to the \textblue{top-most cell}. In contrast, first-order parsers make these two decisions jointly by averaging all the cells that \textred{cross the left and right children} to unify the representation of lexicon and syntax.}
    \label{fig:chart}
\end{figure}

\subsection{First-order Constituency Parsing}
\label{sec:first_cky}

Efficient computing requires batchifying the inside pass of the CKY algorithm for parallel dynamic programming on GPUs \cite{stern-etal-2017-minimal,ijcai2020p0560}. Within the CKY framework, \citet{wang-utiyama-cont-disc} introduce a large tensor as the chart for dynamic programming, where $G{(l,r,\ve{c})}$ refers to the total scores of all trees spanning from $l$ to $r$ with code $\ve{c}$ as the top label, while $g(l,r,\ve{c})$ stands for a single constituent. The algorithm starts from single-word spans and incrementally computes larger spans by enumerating splitting positions and summing children with the top span. \begin{align}
    & G{(l,r,\ve{c})} \gets \sum_{m=l}^{r-1}{\mathchild{G{(l,m,\cdot)}}\ +} \nonumber \\ 
    & \qquad \mathchild{G{(m+1,r,\cdot)}} + \mathold{g(l,r,\ve{c})} \label{eq:zero_cky}
\end{align}

This procedure has been widely employed as a practical standard \cite{ijcai2020p0560,wang-etal-2023-24}. However, we notice that natively using it for unsupervised parsing is not sufficient. As shown in Figure~\ref{fig:chart}, the crux is that even though Equation~\ref{eq:zero_cky} enumerates all valid splitting positions, the span score $g{(l,r,\ve{c})}$ does not take the splitting positions into consideration. According to Equation~\ref{eq:zero_g}, this score depends only on the leftmost and rightmost tokens, regardless of the chosen splitting positions. In other words, different choices of splitting positions do not vary the code scores of top spans. Therefore, performing contrastive hashing by using such scores barely provides any effective information for unsupervised parsing. We refer to this kind of CKY as zero-order CKY.

Naturally, the most intuitive solution is upgrading to first-order CKY by taking the splitting position $m$ into consideration through introducing a novel span score function $g{(l,r,m,\ve{c})}$. \begin{align}
    & G{(l,r,\ve{c})} = \sum_{m=l}^{r-1}{\mathchild{G{(l,m,\cdot)}}\ +} \nonumber \\ 
    & \qquad \mathchild{G{(m+1,r,\cdot)}} + \mathnew{g(l,r,m,\ve{c})} \label{eq:first_cky}
\end{align} And instead of relying only on the leftmost and the rightmost hidden states, we use the averaged representation of the left and right children, respectively. \begin{gather}
    g(l,r,m,\ve{c}) = \sum_{k=1}^{K}{g_k(l,r,m,c^k)} \label{eq:first_g} \vh \\
    g_k(l,r,m,+1) = \dfrac{(\ma{w}_k^Q\overline{\ve{h}}_{l:m})^{\top}(\ma{w}_k^K\overline{\ve{h}}_{m+1:r})}{\sqrt{d_k}} \nonumber \\
    \overline{\ve{h}}_{l:m} = \mean_{l\le i\le m}{\ve{h}_i} \nonumber \vh \\
    \overline{\ve{h}}_{m+1:r} =\mean_{m< j\le r}{\ve{h}_j} \nonumber
\end{gather} Where $\overline{\ve{h}}_{l:m}$ and $\overline{\ve{h}}_{m+1:r}$ are the averaged representation of the left and right children, respectively. In this way, the splitting position influences the scores of binary codes through children hidden states.

However, naively computing $g_k(l,r,m,+1)$ requires additional computational resources for averaging vectors and performing dot products in real-time, which heavily slows down training and inference. Fortunately, through simple derivation, we note that the new score can be obtained by merely averaging the old scores. Upgrading CKY from zero-order to first-order then introduces almost no additional delay by applying this trick. \begin{gather}
    \mathnew{g_k(l,r,m,+1)} = \mean_{l\le i\le m < j \le r}{~\mathold{g_k(i,j,+1)}} \vh
\end{gather} 

According to this definition, the new scores can be interpreted as being obtained by averaging the left and right children, respectively, and then calculating the scores for construing a span across them. Different choices of splitting positions result in different representations of the left and right children, leading to different bit scores for the top span. Since scores reflect the substructure of spans, aligning and uniformalizing these scores in Hamming space using contrastive learning is equivalent to aligning and uniformalizing the subtrees in syntactic structure space. Hence, our method can also be considered relevant to syntactic distance \cite{shen-etal-2018-straight,shen2018ordered}. Additionally, we also assign a score of $0$ for the $-1$ case. \begin{gather}
    g_k(l,r,m,-1) = 0
\end{gather} And extend the marginal probabilities as well. \begin{gather}
    \mu_k{(l,r,m,c^k)} = \dfrac{\partial\,\log{Z}}{\partial\,g_k(l,r,m,c^k)}
\end{gather}

\subsection{Unsupervised Contrastive Hashing}
\label{sec:unsup_hash}

We define our similarity in a manner similar to \citet{wang-utiyama-cont-disc}. As we have upgraded the bit-level CKY module from zero-order to first-order, we also upgrade the binarization procedure. \begin{gather}
    \ve{c} = [c^1,\dots,c^K] \in \B^{K} \\
    c^k = \begin{cases}
        +1 & \mu_k{(l,r,m,+1)} > \mu_k{(l,r,m,-1)} \nonumber \\
        -1 & \text{otherwise}
    \end{cases}
\end{gather} and the similarity function as follows. \begin{gather}
    s(i,j) = \dfrac{1}{K}\sum_{k=1}^K{~\mu_k(l_i,r_i,m_i,c_j^k)}
\end{gather}

Unlike in the supervised settings of \citet{wang-utiyama-cont-disc}, we aim to obtain constituency parsers without training them on annotated treebanks, i.e., $\set{\agl{l_i, r_i, y_i}}_{i=1}^{2n-1}$. Therefore, it is difficult for us to constrain the search space as Equation~\ref{eq:sup_argmax} and to divide spans according to ground-truth labels as Equation~\ref{eq:sup_np}. Thus, we unlock these restrictions and let parsers determine constituent boundaries and binary labels jointly through searching in an unconstrained space $\ca{t}{[\cdot,\cdot,\cdot]}$. \begin{gather}
    \prd{\ve{t}}=\set{\agl{\prd{l}_i,\prd{r}_i,\prd{\ve{c}}_i}}_{i=1}^{2n-1} \gets \argmax_{\ve{t}\in\ca{t}{[\cdot,\cdot,\cdot]}}{~p(\ve{t})} \label{eq:unsup_argmax} \vh
\end{gather} After that, since we neither have access to the ground-truth labels $y_i$, we turn to use the lexicons in spans $w_{\hat{l}_i},\dots,w_{\hat{r}_i}$ as the labels to divide these selected spans. In this way, pulling or pushing spans is determined solely on surface textual features. Besides, since a portion of input tokens are masked out during the augmentation stage, our parsers can be considered a masked language model as well, except that they are trained with a contrastive objective at the span level. \begin{equation}
\begin{gathered}
    \ca{p}=\set{j\mid w_{\prd{l}_i:\prd{r}_i}= w_{\prd{l}_j:\prd{r}_j}} \\
    \ca{n}=\set{j\mid w_{\prd{l}_i:\prd{r}_i}\not= w_{\prd{l}_j:\prd{r}_j}} 
\end{gathered}
\label{eq:unsup_np}
\end{equation}

From the perspective of training, as \citet{wang-etal-2023-24} mentioned, one of the most appealing properties of contrastive learning is that it can convert tasks from \emph{wh-questions} to \emph{yes-no questions}. Conventional classification approaches demand embedding vectors for all spans, but enumerating them all is clearly intractable. According to Appendix~\ref{sec:stats}, we note that even disregarding the sparsity, employing an embedding with millions of entries is barely practical due to its huge memory consumption. In contrast, our contrastive hashing only needs to know if spans are identical or not, allowing it to pull or push their representations directly without needing to introduce specific embeddings. This property makes previously intractable training feasible and efficient.

From the perspective of representation learning, contrastive learning aims to maximize the distinguishability between instances. In our model, this corresponds to maximizing the distinguishability between subtrees. For parsing, choosing the splitting positions that minimize the internal differences within subtrees is equivalent to maximizing the differences across subtrees. In other words, parsing can be considered as a procedure of searching the minimum entropy tree formed by repeatedly merging the most similar contiguous subtrees, thus, it aligns with the learning objective of contrastive hashing. We believe this explains why such a contrastive hashing procedure results in syntactic trees rather than other structures, and this provides justification for our use of contrastive learning.

\subsection{Instance Selection}
\label{sec:loss}

Contrastive learning \cite{gao-etal-2021-simcse} learns informative representation through pulling together positive and pushing apart negative instances. \citet{wang-utiyama-cont-disc} enumerate each instance $i$ and compare it with all instances in the batch $j\in\hat{\ve{t}}$ to compute the instance-level loss $\ell{(i,\mu,\hat{\ve{t}})}$, and then aggregate all these losses as the batch-level loss $\loss$. By using $\log\sum\exp$ as a approximation of $\max$, \begin{equation}
\max_{x\in\ca{x}}{~(x)} \approx \log{\sum_{x\in\ca{x}}\exp{(x)}} \label{eq:max}
\end{equation} They tweaked those commonly used contrastive objectives into unified formats as follows, where $\ca{s}=\set{i}$ is simply defined as the instance itself. \begin{align}
    \ell_{\operatorname{self}} &\approx \mathneg{\max_{\ca{n}\cup\ca{p}}{s(i,j)}} - \mathpos{s(i,i)} \\
    \ell_{\operatorname{sup}} &\approx \mathneg{\max_{\ca{n}\cup\ca{p}}{s(i,j)}} - \mathpos{\mean_{\ca{p}}{s(i,j)}} \\
    \ell_{\operatorname{hash}} &\approx \mathneg{\max_{\ca{n}\cup\ca{s}}{s(i,j)}} - \mathpos{s(i,i)} \\
    \ell_{\max} &\approx \mathneg{\max_{\ca{n}\cup\ca{s}}{s(i,j)}} - \mathpos{\max_{\ca{p}}{s(i,j)}} \\
    = \max&{\left(\max_{\ca{n}}{s(i,j)},\mathbalance{s(i,i)}\right)} - \mathpos{\max_{\ca{p}}{s(i,j)}} \nonumber
\end{align} Objective function $\ell_{\operatorname{self}}$ is commonly utilized in scenarios involving only a single positive instance. \citet{NEURIPS2020_d89a66c7} then extended it as $\ell_{\operatorname{sup}}$ to handle multiple positive instances scenarios, and it was later surpassed by $\ell_{\operatorname{hash}}$ and $\ell_{\max}$. For more details, we refer readers to their original papers \cite{wang-etal-2023-24,wang-utiyama-cont-disc}.

Briefly speaking, objective $\ell_{\operatorname{hash}}$ assumes there is only one true positive and excludes potential false negatives and positives from both terms, with $\ca{p}$ replaced with $\ca{s}$. Moreover, $\ell_{\max}$ adopts a different approach to handling multiple positive instances. They still assume there is only one true positive instance among $\ca{p}$, but they dynamically select the closest one as the true positive, instead of statically selecting $\ca{s}$. By imposing such a weak alignment signal, they also avoid the geometric center issue of $\ell_{\operatorname{sup}}$. However, we found that for tasks with large label vocabularies, such as language models, this signal turns out to be too weak. Therefore, instead of pulling only the closest pairs, we propose to mainly focus on the farthest pairs, \begin{gather}
    \ell_{\min} \approx \mathneg{\max_{\ca{n}\cup\ca{s}}{s(i,j)}} - \mathpos{\min_{\ca{p}}{s(i,j)}} \\
    = \max{\left(\max_{\ca{n}}{s(i,j)},\mathbalance{s(i,i)}\right)} - \mathpos{\min_{\ca{p}}{s(i,j)}} \nonumber
\end{gather} by introducing a differentiable approximation of $\min$ operator in a simialar matter to Equation~\ref{eq:max}. \begin{gather}
    \min_{x\in\ca{x}}{~(x)} \approx -\log{\sum_{x\in\ca{x}}\exp{(-x)}}
\end{gather}

According to experimental results of previous work, excluding potential false negatives seems to be an effective solution, and it also balances the two terms well. However, since $\ell_{\min}$ introduces a strong alignment signal in the positive term, this balance is disrupted. We have consistently observed that the $\ell_{\min}$ model suddenly collapses and starts returning only trivial right-branching trees. We hypothesize the reason is that there is no corresponding uniformity signal in the negative term to balance this strong alignment signal. However, naively adding $\min_{\ca{p}}$ to the negative term as the balancing term leads to a new issue, \begin{equation}
    \max{\left(\max_{\ca{n}}{s(i,j)},\mathgreen{\min_{\ca{p}}{s(i,j)}}\right)} - \mathpos{\min_{\ca{p}}{s(i,j)}} \nonumber
\end{equation} that is when the number of positives is large, positives $\ca{p}$ dominate the gradients, leaving insufficient supervision signals to the true negatives $\ca{n}$. Therefore, we propose to limit the total gradients of positives to be the same magnitude as single positive by introducing another approximation of $\min$. \begin{equation} 
    \mmin_{x\in\ca{x}}{~(x)} \approx -\log{\left(\dfrac{1}{\abs{\ca{x}}}\sum_{x\in\ca{x}}\exp{(-x)}\right)} 
\end{equation} In this way, we propose $\ell_{\mmin}$ as a balanced version, with ${}_{\min{\ca{p}}}$ in the negative term replaced by ${}_{\mmin{\ca{p}}}$. \begin{gather}
    \ell_{\mmin} \approx \mathneg{\max_{\ca{n}\cup\set{\mmin_{\ca{p}}}}{s(i,j)}} - \mathpos{\min_{\ca{p}}{s(i,j)}} \\
    = \max{\left(\max_{\ca{n}}{s(i,j)},\mathgreen{\mmin_{\ca{p}}{s(i,j)}}\right)} - \mathpos{\min_{\ca{p}}{s(i,j)}} \nonumber
\end{gather} 

\subsection{Architecture}
\label{sec:arch}

Following \citet{wang-utiyama-cont-disc}, our model also consists of a pre-trained language model, an attention hash layer, and a bit-level CKY module. The only difference is that we upgrad CKY from zero-order to first-order, which enhances its ability to unify the representation of lexicon and syntax.

Although it is also a masked language model, our model does not require introducing a large embedding matrix for calculating token classification in the output layer. Since it relies on the attention hash layer to produce binary codes of spans, the number of parameters in the output layer is reduced from $\abs{\ca{y}}\times d$ to two $K\times \lceil \dfrac{d}{K} \rceil\times d$.

\subsection{Training and Inference}
\label{sec:train_infer}

During the training stage, sentences are fed into the model twice to obtain two different views by being augmented with different dropout masks. We calculate the marginal probabilities $\mu{}^1$ and $\mu{}^2$, and then predict constituency trees $\prd{\ve{t}}{}^1$ and $\prd{\ve{t}}{}^2$ on these two versions, respectively. For each view, we select the corresponding span scores from the marginal probabilities of one view, according to the predicted tree of the other view, and then perform span-level contrastive hashing by using the objectives above and average them as the batch loss. \begin{gather}
    \loss = \mean_{i\in\prd{\ve{t}}{}^{2}}{~\ell{(i,\mu{}^{1},\prd{\ve{t}}{}^{2})}} + \mean_{i\in\prd{\ve{t}}{}^{1}}{~\ell{(i,\mu{}^{2},\prd{\ve{t}}{}^{1})}}
\end{gather}

Since unsupervised constituency parsing only aims at detecting the span boundaries without needing to predict labels, we do not need to build the code vocabulary as \citet{wang-utiyama-cont-disc} did. During the inference stage, we simply search for the most probable constituency parsing trees in an unconstrained space with the Cocke-Kasami-Younger (CKY) algorithm \cite{kasami1966efficient}. 

\section{Experiments}
\label{sec:exp}

\begin{table}[!t]
    \centering
    \adjustbox{width=\columnwidth}{
    \begin{tabular}{lcc}
    \toprule
    \multicolumn{1}{c}{\multirow{2}{*}{\textsc{Model}}} & \multicolumn{2}{c}{\textsc{Ptb}} \\
    & \small{\textsc{Mean}} & \small{\textsc{Max}} \\
    \midrule
    PRPN \cite{shen2018neural}\im & 37.4 & 38.1 \\
    URNNG \cite{kim-etal-2019-unsupervised}\im & - & 45.4 \\
    ON-LSTM \cite{shen2018ordered}\im & 47.7 & 49.4 \\
    R2D2 \cite{hu-etal-2021-r2d2}\im & 48.1 & - \\
    Fast R2D2 \cite{hu-etal-2022-fast}\im & 48.9 & - \\
    StructFormer \cite{shen-etal-2021-structformer}\im & 54.0 & - \\
    C-PCFG \cite{kim-etal-2019-compound}\ex & 55.2 & 60.1 \\
    NL-PCFG \cite{zhu-etal-2020-return}\ex & 55.3 & - \\
    DIORA \cite{drozdov-etal-2019-unsupervised-latent}\im & 55.7 & 56.2 \\
    GPST \cite{hu-etal-2024-generative}\im & 57.5 & - \\
    S-DIORA \cite{drozdov-etal-2020-unsupervised}\im & 57.6 & 63.9 \\
    TN-PCFG \cite{yang-etal-2021-pcfgs}\ex & 57.7 & 61.4 \\
    NBL-PCFG \cite{yang-etal-2021-neural}\ex & 60.4 & - \\
    CT \cite{cao-etal-2020-unsupervised}\prob & 62.8 & 65.9 \\
    Co \cite{maveli-cohen-2022-co}\prob & 63.1 & 66.8 \\
    Rank-PCFG \cite{yang-etal-2022-dynamic}\ex & 64.1 & - \\
    ReCAT \cite{hu2024augmenting}\im & \sotb{65.0} & - \\
    SN-PCFG \cite{liu-etal-2023-simple}\ex & \sota{65.1} & - \\
    \midrule
    {\small \textsc{For Reference}} \\
    Ensemble \cite{shayegh2024ensemble} & 70.4 & 71.9 \\
    Left Branching & 8.7 & 8.7 \\
    Right Branching & 39.5 & 39.5 \\
    Oracle & 84.3 & 84.3 \\
    \midrule
    Ours\im\ ({\small\textsc{Bert}${}_{\textsc{Base}}$} - 16 bits) & 55.3 & 58.8 \\
    Ours\im\ ({\small\textsc{Bert}${}_{\textsc{Base}}$} - 20 bits) & \sotb{56.7} & 59.8 \\
    Ours\im\ ({\small\textsc{Bert}${}_{\textsc{Base}}$} - 24 bits) & \sota{57.4} & 59.6 \\
    Ours\im\ ({\small\textsc{Bert}${}_{\textsc{Base}}$} - 28 bits) & 54.5 & 60.9\\
    \midrule
    Ours\im\ ({\small\textsc{RoBERTa}${}_{\textsc{Base}}$} - 8 bits) & 56.5 & 63.1 \\
    Ours\im\ ({\small\textsc{RoBERTa}${}_{\textsc{Base}}$} - 12 bits) & 58.0 & 62.9 \\
    Ours\im\ ({\small\textsc{RoBERTa}${}_{\textsc{Base}}$} - 16 bits) & \sota{62.4} & 64.1 \\
    Ours\im\ ({\small\textsc{RoBERTa}${}_{\textsc{Base}}$} - 20 bits) & \sotb{59.6} & 63.9 \\
    \bottomrule
    \end{tabular}
    }
    \caption{Experiments of unsupervised constituency parsing on the PTB dataset. The columns {\small${\textsc{Mean}}$} and {\small${\textsc{Max}}$} display the averaged and the maximal unlabeled sentence-level F$_1$ scores. The \sota{bold numbers} and the \sotb{underlined numbers} indicate the best and the second-best performance. $\rawim\rawex\rawprob$ stands for implicit grammar, explicit grammar, and probing methods, respectively.}
    \label{tab:ptb}
\end{table}

\begin{table}[!t]
    \centering
    \adjustbox{width=\columnwidth}{
    \begin{tabular}{lcc}
    \toprule
    \multicolumn{1}{c}{\multirow{2}{*}{\textsc{Model}}} & \multicolumn{2}{c}{\textsc{Ctb}} \\
    & \small{\textsc{Mean}} & \small{\textsc{Max}} \\
    \midrule
    ON-LSTM \cite{shen2018ordered}\im & 25.4 & 25.7 \\
    PRPN \cite{shen2018neural}\im & 30.4 & 31.5 \\
    Rank-PCFG \cite{yang-etal-2022-dynamic}\ex & 32.4 & - \\
    C-PCFG \cite{kim-etal-2019-compound}\ex & 36.0 & 39.8 \\
    TN-PCFG \cite{yang-etal-2021-pcfgs}\ex & 39.2 & - \\
    Co \cite{maveli-cohen-2022-co}\prob & 41.8 & 43.3 \\
    SC-PCFG \cite{liu-etal-2023-simple}\ex & 42.9 & - \\
    R2D2 \cite{hu-etal-2021-r2d2}\im & \sotb{44.9} & - \\
    Fast R2D2 \cite{hu-etal-2022-fast}\im & \sota{45.3} & - \\
    \midrule
    {\small \textsc{For Reference}} \\
    Left Branching & 9.7 & 9.7 \\
    Right Branching & 20.0 & 20.0 \\
    Oracle & 81.1 & 81.1 \\
    \midrule
    Ours\im\ ({\small\textsc{Bert}${}_{\textsc{Base}}$} - 28 bits) & 41.2 & 49.0 \\
    Ours\im\ ({\small\textsc{Bert}${}_{\textsc{Base}}$} - 32 bits) & 43.1 & 49.5 \\
    Ours\im\ ({\small\textsc{Bert}${}_{\textsc{Base}}$} - 36 bits) & \sota{47.1} & 49.6 \\
    Ours\im\ ({\small\textsc{Bert}${}_{\textsc{Base}}$} - 40 bits) & \sotb{43.6} & 49.5 \\
    \midrule
    Ours\im\ ({\small\textsc{RoBERTa}${}_{\textsc{Base}}$} - 36 bits) & 46.4 & 50.2 \\
    Ours\im\ ({\small\textsc{RoBERTa}${}_{\textsc{Base}}$} - 40 bits) & 45.4 & 50.0 \\
    Ours\im\ ({\small\textsc{RoBERTa}${}_{\textsc{Base}}$} - 44 bits) & \sota{48.5} & 49.6 \\
    Ours\im\ ({\small\textsc{RoBERTa}${}_{\textsc{Base}}$} - 48 bits) & \sotb{47.0} & 50.3 \\
    \bottomrule
    \end{tabular}
    }
    \caption{Experiments of unsupervised constituency parsing on the CTB dataset.}
    \label{tab:ctb}
\end{table}

\subsection{Settings}
\label{sec:settings}

Experiments are conducted on the commonly used datasets Penn Treebank (PTB) \cite{marcus-etal-1993-building} and Chinese Treebank 5.1 (CTB) \cite{XUE_XIA_CHIOU_PALMER_2005}.

Following previous settings \cite{shen2018neural,shen2018ordered,zhao-titov-2021-empirical}, we use the same preprocessing pipeline to discard punctuation in all splits. Although this pipeline may not be the best choice for pre-trained language models and might result in some information loss, since language models are commonly trained with punctuated corpora, we follow this setting only to provide comparable results to previous work. Regarding the evaluation metric, we follow \citet{kim-etal-2019-compound} to remove trivial spans, i.e., single-word and entire-sentence spans, calculate unlabeled sentence-level F1 scores, and take the average across all sentences.

We use the deep learning framework \texttt{PyTorch} \cite{NEURIPS2019_bdbca288} to implement our models and download checkpoints of pre-trained languages from \texttt{huggingface/transformers} \cite{wolf-etal-2020-transformers}. Different from some recent work \cite{yang-etal-2022-dynamic,liu-etal-2023-simple}, which require customizing CUDA kernels with \texttt{Triton} \cite{10.1145/3315508.3329973}, our model can be easily and efficiently implemented with pure \texttt{PyTorch}.

We collect sentences until the total number of spans reaches 1024 to keep the contrastive hashing stable, since it is performed at the span level. We use the Adam optimizer \cite{kingma2017adam,loshchilov2018decoupled} and set the number of warmup and training steps to \num{4000} and \num{20000}, respectively. We randomly select a portion of tokens and replace them with \texttt{[MASK]}, following the standard augmentation strategy of masked language models. For PTB experiments, we use checkpoints \texttt{bert-base-cased} \cite{devlin-etal-2019-bert} and \texttt{roberta-base} \cite{DBLP:journals/corr/abs-1907-11692}. For CTB experiments, we use checkpoints \texttt{bert-base-chinese} \cite{devlin-etal-2019-bert} and \texttt{chinese-roberta-wwm-ext} \cite{cui-etal-2020-revisiting}.

We use a single NVIDIA Tesla V100 graphics card to conduct our experiments. Training takes around 30 minutes, which is much faster than the several days of training required by \citet{cao-etal-2020-unsupervised} and \citet{drozdov-etal-2019-unsupervised}. Since we do not modify the architecture of the language model but simply append a hash layer to it, we can fine-tune existing pre-trained language models without needing to train them from scratch, as done by \citet{hu-etal-2022-fast,hu-etal-2024-generative}. For each setting, we run it four times with different random seeds and report the averaged scores in the following tables.

\begin{table}[!t]
    \centering
    \adjustbox{width=\columnwidth}{
    \begin{tabular}{cccccc}
    \toprule
    \multirow{2}{*}{\textblue{\textsc{Neg}}} & \multirow{2}{*}{\textred{\textsc{Pos}}} & \multirow{2}{*}{\textsc{Loss}} & \multicolumn{2}{c}{\textsc{Ptb}} \\
    & & & \small{\textsc{Mean}} & \small{\textsc{Max}} \\
    \midrule
    \multirow{3}{*}{$\max_{\ca{n}\cup\ca{p}}$} & ${}_{\ca{s}}$ & $\ell_{\operatorname{self}}$ & 39.9 & 40.4 \\
     & $\max_{\ca{p}}$ & - & 44.0 & 54.0 \\
     & $\min_{\ca{p}}$ & - & 48.8 & 61.8 \\
    \midrule
    \multirow{3}{*}{$\max_{\ca{n}\cup\ca{s}}$} & ${}_{\ca{s}}$ & $\ell_{\operatorname{hash}}$ & 39.9 & 40.3 \\
     & $\max_{\ca{p}}$ & $\ell_{\max}$ & 45.5 & 50.1 \\
     & $\min_{\ca{p}}$ & $\ell_{\min}$ & \sotb{58.2} & 60.6 \\
    \midrule
    \multirow{3}{*}{$\max_{\ca{n}\cup\{\mmin_{\ca{p}}\}}$} & ${}_{\ca{s}}$ & - & 35.2 & 49.1 \\
     & $\max_{\ca{p}}$ & - & 47.5 & 53.9 \\
     & $\min_{\ca{p}}$ & $\ell_{\mmin}$ & \sota{62.4} & 64.1 \\
    \bottomrule
    \end{tabular}
    }
    \caption{Ablation study of instance selection strategies in constituency parsing experiments. Columns \textsc{Neg} and \textsc{Pos} display the selection strategies for negatives and positives, respectively. \textsc{Loss} shows this combination corresponds to which loss definition.}
    \label{tab:ablation}
\end{table}

\subsection{Main Results}
\label{sec:main}

On the English dataset PTB, as shown in Table~\ref{tab:ptb}, our model reaches its peak performance at 24 bits and 16 bits when using BERT and RoBERTa pre-trained language models, respectively. We consistently surpass all other implicit grammar models. Due to the relatively small size of PTB, the probing methods by \citet{cao-etal-2020-unsupervised} and \citet{maveli-cohen-2022-co} utilized additional text data for training. Even without using such extra data, our model still achieves performance very close to theirs.

Our model outperforms all existing models by a large margin on the Chinese dataset CTB, as shown in Table~\ref{tab:ctb}. Explicit grammar models that perform well on English datasets \cite{yang-etal-2022-dynamic, liu-etal-2023-simple} do not achieve similar success on the Chinese dataset. Additionally, we notice that our model requires much more bits than on the English dataset, i.e., 36 and 44, to reach their full potential. We hypothesize that this is due to the relatively small size of the Chinese dataset, as shown in Appendix~\ref{sec:stats}, which prevents the models from being fully trained to encode lexicon and syntax features within only a few bits.

\subsection{Ablation Studies}
\label{sec:ablation}

Table~\ref{tab:ablation} reveals how the different combinations of negative and positive terms affect performance. First of all, we notice that once $\min_{\ca{p}}$ is employed, regardless of which negative terms are used along with it, the models consistently result in high scores in the {\small${\textsc{Max}}$} column. On the contrary, without employing $\min_{\ca{p}}$, these scores dramatically drop. This confirms our statement that for tasks with large label vocabularies, positive terms require strong alignment signals to learn effective representations. Moreover, when it comes to the {\small${\textsc{Mean}}$} column, whether the term $\max_{\ca{n}\cup\{\mmin_{\ca{p}}\}}$ is employed determines whether the high scores of $\min_{\ca{p}}$ can be maintained. We also notice that $\max_{\ca{n}\cup\ca{s}}$ consistently outperforms $\max_{\ca{n}\cup\ca{p}}$. This indicates that simply pushing away all instances of $\ca{p}$ indeed introduces the false negatives issue. As \citet{wang-etal-2023-24, wang-utiyama-cont-disc} claims, retaining only $\ca{s}$ mitigates this issue, but when $\ca{p}$ is introduced back to the positive term under a strong alignment, the lack of uniformity signals brings a new imbalance issue, and our solution $\ell_{\mmin}$ re-balances them by using $\mmin_{\ca{p}}$ in both terms.

\subsection{Case Studies}
\label{sec:case}

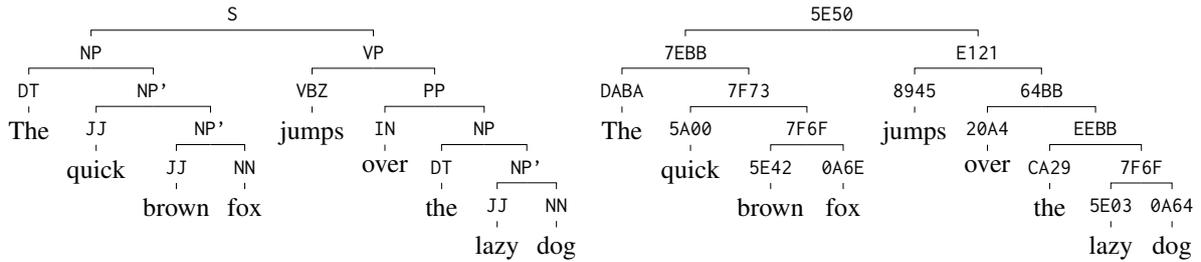
\begin{figure*}[t]
\centering
\adjustbox{width=\textwidth}{
\begin{tikzpicture}
\centering
\Tree [.\code{S}
        [.\code{NP}
          [.\code{DT} The ]
          [.\code{NP'} [.\code{JJ} quick ] [.\code{NP'} [.\code{JJ} brown ] [.\code{NN} fox ] ] ] ]
        [.\code{VP}
          [.\code{VBZ} jumps ]
          [.\code{PP}
            [.\code{IN} over ]
            [.\code{NP} [.\code{DT} the ] [.\code{NP'} [.\code{JJ} lazy ] [.\code{NN} dog ] ] ] ] ] ]
\end{tikzpicture}
\begin{tikzpicture}
\centering
\Tree [.\code{5E50}
        [.\code{7EBB}
          [.\code{DABA} The ]
          [.\code{7F73}
            [.\code{5A00} quick ]
            [.\code{7F6F} [.\code{5E42} brown ] [.\code{0A6E} fox ] ] ] ]
        [.\code{E121}
          [.\code{8945} jumps ]
          [.\code{64BB}
            [.\code{20A4} over ]
            [.\code{EEBB}
              [.\code{CA29} the ]
              [.\code{7F6F}
                [.\code{5E03} lazy ]
                [.\code{0A64} dog ] ] ] ] ] ]
\end{tikzpicture}
}
\caption{Derivation of the sentence \emph{The quick brown fox jumps over the lazy dog}. The left side is the ground-truth consistency tree, and the right side is our parsing result with binary labels represented in \texttt{hexadecimal} format.}
\label{fig:example}
\end{figure*}

Figure~\ref{fig:example} shows an example of our parsing results, with more examples available in Appendix~\ref{sec:example}. Relying on the implicitly induced grammar, our model provides remarkably accurate parsing results, with all constituents correctly selected. Additionally, the hashing results also demonstrate the impressive capability in discovering syntactic categories. For instance, both preterminal symbols like adjectives, e.g., \emph{quick} ({\code{5A00}}), \emph{brown} ({\code{5E42}}), and \emph{lazy} ({\code{5E03}}), and nonterminal symbols like noun phrases, e.g., \emph{the quick brown fox} ({\code{7EBB}}) and \emph{the lazy dog} ({\code{EEBB}}), are assigned similar and relevant binary codes to each other. This phenomenon can also be observed in sentences in Appendix~\ref{sec:example}, indicating that our parser can accurately reveal both part-of-speech and constituent features at different hierarchical levels.

\section{Related Work} 
\label{sec:related}

Syntactic language models, as a historical and important field of language models, had been widely studied even before the deep learning era \cite{CHELBA2000283,charniak-2001-immediate,roark-2001-probabilistic,klein-manning-2002-generative,klein-manning-2004-corpus,bod-2006-subtrees,bod-2006-unsupervised}. After that, \citet{shen-etal-2018-straight,shen2018neural,shen2018ordered} added syntactic inductive bias to LSTM by introducing master gates to control the information flow in hierarchical directions, thereby enabling the model to learn syntactic distance. Under this framework, by training recurrent language models in the usual way, they can obtain parsers that implicitly structure sentences according to the learned syntactic distances. They have also successfully applied this method to transformers \cite{shen-etal-2021-structformer}. 

Implicit grammar models induce grammar during the training process by incrementally constructing larger span representations. \citet{kim-etal-2019-unsupervised} were the first to extend the recurrent neural network grammar (RNNG) \cite{dyer-etal-2016-recurrent} from supervised to unsupervised settings. They build parsing trees through continuously shifting and reducing, without introducing explicit production rules. Additionally, DIORA \cite{drozdov-etal-2019-unsupervised-latent,drozdov-etal-2020-unsupervised} construct span representations and update charts in both inside and outside passes, and then encourage consistency between them. Similarly, R2D2 \cite{hu-etal-2021-r2d2,hu-etal-2022-fast} is trained in a similar manner, but with Gumbel-softmax \cite{jang2017categorical} introduced during the tree construction.

Apart from implicit grammar models, explicitly inducing probabilistic context-free grammar (PCFG) is also widely focused. \citet{kim-etal-2019-compound} first brought PCFG approaches back with a neural parameterization technique and trained language models to reconstruct entire input sentences token by token. \citet{zhu-etal-2020-return} shows that additionally modeling lexical dependencies \cite{collins-2003-head} is effective, and \citet{jiang-etal-2016-unsupervised,yang-etal-2021-neural} further confirmed extending to bilexical dependencies is even more beneficial. Besides, \citet{yang-etal-2021-pcfgs,yang-etal-2022-dynamic,liu-etal-2023-simple} claimed that the limited number of symbols is a bottleneck of PCFG induction, and proposed to introduce more symbols by applying tensor decomposition to overcome the cubic computational complexity.

How much syntactic knowledge is preserved within the ordinary language model is also a question worth considering. \citet{marecek-rosa-2018-extracting,marecek-rosa-2019-balustrades} noticed the value of attention scores first. They defined distance functions similar to our Equation~\ref{eq:first_g} for probing. However, they did not consider splitting positions, and relied only on fixed attention heads without having them fine-tuned. As a result, their methods resulted in limited accuracy. \citet{hewitt-manning-2019-structural,wang-etal-2019-tree,Kim2020Are,li-etal-2020-heads,bai-etal-2021-syntax} then introduced parameterized functions to prob syntactic distances on hidden states and attention scores, and then fine-tuned the entire models. \citet{cao-etal-2020-unsupervised,maveli-cohen-2022-co} fine-tuned pre-trained language models with an additional classifier to distinguish manually generated constituents and distituents, and utilized predictions from this classifier to determine splitting positions on parsing trees during the evaluation stage.

In various senses, our approach confirmed the conclusions of many previous works and further pushed their limits. First, by switching the language models from token-level to span-level, we confirmed that modeling lexical dependencies is beneficial and extending this modeling to all tokens in children is more effective. Additionally, by introducing binary representation, we confirmed that employing more symbols is advantageous and further scaling up to $2^K$ can help parsers do further better. Third, we confirmed that constructing span representations and updating the chart is helpful, and unifying the representation of lexicon and syntax leads to more competitive results. Finally, we confirmed the multi-head attention scores already preserve syntactic information, and fine-tuning them can help probe for more insightful features.

\section{Conclusions}
\label{sec:conclusions}

In this paper, we confirmed that the information-preserving capability of binary representation is effective at both lexicon and syntax levels, and we demonstrated that it is feasible to elicit parsers from pre-trained language models by leveraging this capability. We achieved this by upgrading bit-level CKY from zero-order to first-order, extending contrastive hashing from supervised to unsupervised, and proposing a novel objective function to impose stronger yet balanced alignment signals. Experiments show our model achieves competitive performance, and also indicate that the technique for acquiring high-quality syntactic annotations at low cost has now reached a practical stage.

\section*{Limitations}
\label{sec:limitation}

We successfully obtain parsers in an unsupervised manner. Nonetheless, the number of bits remains a hyperparameter that needs to be tuned by testing them individually. Although, in practice, enumerating from 8 to 48 is sufficient for most cases, the relationship between the required number of bits and the specified task remains unclear. Therefore, we aim to explore this issue in future work. Moreover, we simply define the left and right span representations as the average of their token vectors, as shown in Equation~\ref{eq:first_g}. The reason for using such a naive definition is a compromise for the sake of speed. However, it is evident that this simple linear mapping may not efficiently preserve high-order information, and future work could explore more complex mechanisms.

\bibliography{reference}

\begin{thebibliography}{64}
\providecommand{\natexlab}[1]{#1}

\bibitem[{Bai et~al.(2021)Bai, Wang, Chen, Yang, Bai, Yu, and
  Tong}]{bai-etal-2021-syntax}
Jiangang Bai, Yujing Wang, Yiren Chen, Yaming Yang, Jing Bai, Jing Yu, and
  Yunhai Tong. 2021.
\newblock \href {https://doi.org/10.18653/v1/2021.eacl-main.262}
  {Syntax-{BERT}: Improving pre-trained transformers with syntax trees}.
\newblock In \emph{Proceedings of the 16th Conference of the European Chapter
  of the Association for Computational Linguistics: Main Volume}, pages
  3011--3020, Online. Association for Computational Linguistics.

\bibitem[{Bod(2006{\natexlab{a}})}]{bod-2006-subtrees}
Rens Bod. 2006{\natexlab{a}}.
\newblock \href {https://doi.org/10.3115/1220175.1220284} {An all-subtrees
  approach to unsupervised parsing}.
\newblock In \emph{Proceedings of the 21st International Conference on
  Computational Linguistics and 44th Annual Meeting of the Association for
  Computational Linguistics}, pages 865--872, Sydney, Australia. Association
  for Computational Linguistics.

\bibitem[{Bod(2006{\natexlab{b}})}]{bod-2006-unsupervised}
Rens Bod. 2006{\natexlab{b}}.
\newblock \href {https://aclanthology.org/W06-2912} {Unsupervised parsing with
  {U}-{DOP}}.
\newblock In \emph{Proceedings of the Tenth Conference on Computational Natural
  Language Learning ({C}o{NLL}-X)}, pages 85--92, New York City. Association
  for Computational Linguistics.

\bibitem[{Cao et~al.(2020)Cao, Kitaev, and Klein}]{cao-etal-2020-unsupervised}
Steven Cao, Nikita Kitaev, and Dan Klein. 2020.
\newblock \href {https://doi.org/10.18653/v1/2020.emnlp-main.389} {Unsupervised
  parsing via constituency tests}.
\newblock In \emph{Proceedings of the 2020 Conference on Empirical Methods in
  Natural Language Processing (EMNLP)}, pages 4798--4808, Online. Association
  for Computational Linguistics.

\bibitem[{Charniak(2001)}]{charniak-2001-immediate}
Eugene Charniak. 2001.
\newblock \href {https://doi.org/10.3115/1073012.1073029} {Immediate-head
  parsing for language models}.
\newblock In \emph{Proceedings of the 39th Annual Meeting of the Association
  for Computational Linguistics}, pages 124--131, Toulouse, France. Association
  for Computational Linguistics.

\bibitem[{Chelba and Jelinek(2000)}]{CHELBA2000283}
Ciprian Chelba and Frederick Jelinek. 2000.
\newblock \href {https://doi.org/10.1006/csla.2000.0147} {Structured language
  modeling}.
\newblock \emph{Computer Speech \& Language}, 14(4):283--332.

\bibitem[{Collins(2003)}]{collins-2003-head}
Michael Collins. 2003.
\newblock \href {https://doi.org/10.1162/089120103322753356} {Head-driven
  statistical models for natural language parsing}.
\newblock \emph{Computational Linguistics}, 29(4):589--637.

\bibitem[{Cui et~al.(2020)Cui, Che, Liu, Qin, Wang, and
  Hu}]{cui-etal-2020-revisiting}
Yiming Cui, Wanxiang Che, Ting Liu, Bing Qin, Shijin Wang, and Guoping Hu.
  2020.
\newblock \href {https://doi.org/10.18653/v1/2020.findings-emnlp.58}
  {Revisiting pre-trained models for {C}hinese natural language processing}.
\newblock In \emph{Findings of the Association for Computational Linguistics:
  EMNLP 2020}, pages 657--668, Online. Association for Computational
  Linguistics.

\bibitem[{Devlin et~al.(2019)Devlin, Chang, Lee, and
  Toutanova}]{devlin-etal-2019-bert}
Jacob Devlin, Ming-Wei Chang, Kenton Lee, and Kristina Toutanova. 2019.
\newblock \href {https://doi.org/10.18653/v1/N19-1423} {{BERT}: Pre-training of
  deep bidirectional transformers for language understanding}.
\newblock In \emph{Proceedings of the 2019 Conference of the North {A}merican
  Chapter of the Association for Computational Linguistics: Human Language
  Technologies, Volume 1 (Long and Short Papers)}, pages 4171--4186,
  Minneapolis, Minnesota. Association for Computational Linguistics.

\bibitem[{Drozdov et~al.(2020)Drozdov, Rongali, Chen, O{'}Gorman, Iyyer, and
  McCallum}]{drozdov-etal-2020-unsupervised}
Andrew Drozdov, Subendhu Rongali, Yi-Pei Chen, Tim O{'}Gorman, Mohit Iyyer, and
  Andrew McCallum. 2020.
\newblock \href {https://doi.org/10.18653/v1/2020.emnlp-main.392} {Unsupervised
  parsing with {S}-{DIORA}: Single tree encoding for deep inside-outside
  recursive autoencoders}.
\newblock In \emph{Proceedings of the 2020 Conference on Empirical Methods in
  Natural Language Processing (EMNLP)}, pages 4832--4845, Online. Association
  for Computational Linguistics.

\bibitem[{Drozdov et~al.(2019{\natexlab{a}})Drozdov, Verga, Chen, Iyyer, and
  McCallum}]{drozdov-etal-2019-unsupervised}
Andrew Drozdov, Patrick Verga, Yi-Pei Chen, Mohit Iyyer, and Andrew McCallum.
  2019{\natexlab{a}}.
\newblock \href {https://doi.org/10.18653/v1/D19-1161} {Unsupervised labeled
  parsing with deep inside-outside recursive autoencoders}.
\newblock In \emph{Proceedings of the 2019 Conference on Empirical Methods in
  Natural Language Processing and the 9th International Joint Conference on
  Natural Language Processing (EMNLP-IJCNLP)}, pages 1507--1512, Hong Kong,
  China. Association for Computational Linguistics.

\bibitem[{Drozdov et~al.(2019{\natexlab{b}})Drozdov, Verga, Yadav, Iyyer, and
  McCallum}]{drozdov-etal-2019-unsupervised-latent}
Andrew Drozdov, Patrick Verga, Mohit Yadav, Mohit Iyyer, and Andrew McCallum.
  2019{\natexlab{b}}.
\newblock \href {https://doi.org/10.18653/v1/N19-1116} {Unsupervised latent
  tree induction with deep inside-outside recursive auto-encoders}.
\newblock In \emph{Proceedings of the 2019 Conference of the North {A}merican
  Chapter of the Association for Computational Linguistics: Human Language
  Technologies, Volume 1 (Long and Short Papers)}, pages 1129--1141,
  Minneapolis, Minnesota. Association for Computational Linguistics.

\bibitem[{Dyer et~al.(2016)Dyer, Kuncoro, Ballesteros, and
  Smith}]{dyer-etal-2016-recurrent}
Chris Dyer, Adhiguna Kuncoro, Miguel Ballesteros, and Noah~A. Smith. 2016.
\newblock \href {https://doi.org/10.18653/v1/N16-1024} {Recurrent neural
  network grammars}.
\newblock In \emph{Proceedings of the 2016 Conference of the North {A}merican
  Chapter of the Association for Computational Linguistics: Human Language
  Technologies}, pages 199--209, San Diego, California. Association for
  Computational Linguistics.

\bibitem[{Eisner(2016)}]{eisner-2016-inside}
Jason Eisner. 2016.
\newblock \href {https://doi.org/10.18653/v1/W16-5901} {Inside-outside and
  forward-backward algorithms are just backprop (tutorial paper)}.
\newblock In \emph{Proceedings of the Workshop on Structured Prediction for
  {NLP}}, pages 1--17, Austin, TX. Association for Computational Linguistics.

\bibitem[{Gao et~al.(2021)Gao, Yao, and Chen}]{gao-etal-2021-simcse}
Tianyu Gao, Xingcheng Yao, and Danqi Chen. 2021.
\newblock \href {https://doi.org/10.18653/v1/2021.emnlp-main.552} {{S}im{CSE}:
  Simple contrastive learning of sentence embeddings}.
\newblock In \emph{Proceedings of the 2021 Conference on Empirical Methods in
  Natural Language Processing}, pages 6894--6910, Online and Punta Cana,
  Dominican Republic. Association for Computational Linguistics.

\bibitem[{Harris(1954)}]{doi:10.1080/00437956.1954.11659520}
Zellig~S. Harris. 1954.
\newblock \href {https://doi.org/10.1080/00437956.1954.11659520}
  {Distributional structure}.
\newblock \emph{WORD}, 10(2-3):146--162.

\bibitem[{Hewitt and Manning(2019)}]{hewitt-manning-2019-structural}
John Hewitt and Christopher~D. Manning. 2019.
\newblock \href {https://doi.org/10.18653/v1/N19-1419} {{A} structural probe
  for finding syntax in word representations}.
\newblock In \emph{Proceedings of the 2019 Conference of the North {A}merican
  Chapter of the Association for Computational Linguistics: Human Language
  Technologies, Volume 1 (Long and Short Papers)}, pages 4129--4138,
  Minneapolis, Minnesota. Association for Computational Linguistics.

\bibitem[{Hu et~al.(2024{\natexlab{a}})Hu, Ji, Zhu, Wu, and
  Tu}]{hu-etal-2024-generative}
Xiang Hu, Pengyu Ji, Qingyang Zhu, Wei Wu, and Kewei Tu. 2024{\natexlab{a}}.
\newblock \href {https://aclanthology.org/2024.acl-long.145} {Generative
  pretrained structured transformers: Unsupervised syntactic language models at
  scale}.
\newblock In \emph{Proceedings of the 62nd Annual Meeting of the Association
  for Computational Linguistics (Volume 1: Long Papers)}, pages 2640--2657,
  Bangkok, Thailand. Association for Computational Linguistics.

\bibitem[{Hu et~al.(2022)Hu, Mi, Li, and de~Melo}]{hu-etal-2022-fast}
Xiang Hu, Haitao Mi, Liang Li, and Gerard de~Melo. 2022.
\newblock \href {https://aclanthology.org/2022.emnlp-main.181} {Fast-{R}2{D}2:
  A pretrained recursive neural network based on pruned {CKY} for grammar
  induction and text representation}.
\newblock In \emph{Proceedings of the 2022 Conference on Empirical Methods in
  Natural Language Processing}, pages 2809--2821, Abu Dhabi, United Arab
  Emirates. Association for Computational Linguistics.

\bibitem[{Hu et~al.(2021)Hu, Mi, Wen, Wang, Su, Zheng, and
  de~Melo}]{hu-etal-2021-r2d2}
Xiang Hu, Haitao Mi, Zujie Wen, Yafang Wang, Yi~Su, Jing Zheng, and Gerard
  de~Melo. 2021.
\newblock \href {https://doi.org/10.18653/v1/2021.acl-long.379} {{R}2{D}2:
  Recursive transformer based on differentiable tree for interpretable
  hierarchical language modeling}.
\newblock In \emph{Proceedings of the 59th Annual Meeting of the Association
  for Computational Linguistics and the 11th International Joint Conference on
  Natural Language Processing (Volume 1: Long Papers)}, pages 4897--4908,
  Online. Association for Computational Linguistics.

\bibitem[{Hu et~al.(2024{\natexlab{b}})Hu, Zhu, Tu, and Wu}]{hu2024augmenting}
Xiang Hu, Qingyang Zhu, Kewei Tu, and Wei Wu. 2024{\natexlab{b}}.
\newblock \href {https://openreview.net/forum?id=u859gX7ADC} {Augmenting
  transformers with recursively composed multi-grained representations}.
\newblock In \emph{The Twelfth International Conference on Learning
  Representations}.

\bibitem[{Jang et~al.(2017)Jang, Gu, and Poole}]{jang2017categorical}
Eric Jang, Shixiang Gu, and Ben Poole. 2017.
\newblock \href {https://openreview.net/forum?id=rkE3y85ee} {Categorical
  reparameterization with gumbel-softmax}.
\newblock In \emph{International Conference on Learning Representations}.

\bibitem[{Jiang et~al.(2016)Jiang, Han, and Tu}]{jiang-etal-2016-unsupervised}
Yong Jiang, Wenjuan Han, and Kewei Tu. 2016.
\newblock \href {https://doi.org/10.18653/v1/D16-1073} {Unsupervised neural
  dependency parsing}.
\newblock In \emph{Proceedings of the 2016 Conference on Empirical Methods in
  Natural Language Processing}, pages 763--771, Austin, Texas. Association for
  Computational Linguistics.

\bibitem[{Kasami(1966)}]{kasami1966efficient}
Tadao Kasami. 1966.
\newblock \href {https://apps.dtic.mil/sti/citations/AD0631692} {An efficient
  recognition and syntax-analysis algorithm for context-free languages}.
\newblock \emph{Coordinated Science Laboratory Report no. R-257}.

\bibitem[{Khosla et~al.(2020)Khosla, Teterwak, Wang, Sarna, Tian, Isola,
  Maschinot, Liu, and Krishnan}]{NEURIPS2020_d89a66c7}
Prannay Khosla, Piotr Teterwak, Chen Wang, Aaron Sarna, Yonglong Tian, Phillip
  Isola, Aaron Maschinot, Ce~Liu, and Dilip Krishnan. 2020.
\newblock \href
  {https://proceedings.neurips.cc/paper_files/paper/2020/file/d89a66c7c80a29b1bdbab0f2a1a94af8-Paper.pdf}
  {Supervised contrastive learning}.
\newblock In \emph{Advances in Neural Information Processing Systems},
  volume~33, pages 18661--18673. Curran Associates, Inc.

\bibitem[{Kim et~al.(2020)Kim, Choi, Edmiston, and goo Lee}]{Kim2020Are}
Taeuk Kim, Jihun Choi, Daniel Edmiston, and Sang goo Lee. 2020.
\newblock \href {https://openreview.net/forum?id=H1xPR3NtPB} {Are pre-trained
  language models aware of phrases? simple but strong baselines for grammar
  induction}.
\newblock In \emph{International Conference on Learning Representations}.

\bibitem[{Kim et~al.(2019{\natexlab{a}})Kim, Dyer, and
  Rush}]{kim-etal-2019-compound}
Yoon Kim, Chris Dyer, and Alexander Rush. 2019{\natexlab{a}}.
\newblock \href {https://doi.org/10.18653/v1/P19-1228} {Compound probabilistic
  context-free grammars for grammar induction}.
\newblock In \emph{Proceedings of the 57th Annual Meeting of the Association
  for Computational Linguistics}, pages 2369--2385, Florence, Italy.
  Association for Computational Linguistics.

\bibitem[{Kim et~al.(2019{\natexlab{b}})Kim, Rush, Yu, Kuncoro, Dyer, and
  Melis}]{kim-etal-2019-unsupervised}
Yoon Kim, Alexander Rush, Lei Yu, Adhiguna Kuncoro, Chris Dyer, and G{\'a}bor
  Melis. 2019{\natexlab{b}}.
\newblock \href {https://doi.org/10.18653/v1/N19-1114} {Unsupervised recurrent
  neural network grammars}.
\newblock In \emph{Proceedings of the 2019 Conference of the North {A}merican
  Chapter of the Association for Computational Linguistics: Human Language
  Technologies, Volume 1 (Long and Short Papers)}, pages 1105--1117,
  Minneapolis, Minnesota. Association for Computational Linguistics.

\bibitem[{Kingma and Ba(2015)}]{kingma2017adam}
Diederik~P. Kingma and Jimmy Ba. 2015.
\newblock \href {https://arxiv.org/abs/1412.6980} {Adam: A method for
  stochastic optimization}.
\newblock In \emph{The Third International Conference on Learning
  Representations}.

\bibitem[{Kitaev and Klein(2018)}]{kitaev-klein-2018-constituency}
Nikita Kitaev and Dan Klein. 2018.
\newblock \href {https://doi.org/10.18653/v1/P18-1249} {Constituency parsing
  with a self-attentive encoder}.
\newblock In \emph{Proceedings of the 56th Annual Meeting of the Association
  for Computational Linguistics (Volume 1: Long Papers)}, pages 2676--2686,
  Melbourne, Australia. Association for Computational Linguistics.

\bibitem[{Klein and Manning(2004)}]{klein-manning-2004-corpus}
Dan Klein and Christopher Manning. 2004.
\newblock \href {https://doi.org/10.3115/1218955.1219016} {Corpus-based
  induction of syntactic structure: Models of dependency and constituency}.
\newblock In \emph{Proceedings of the 42nd Annual Meeting of the Association
  for Computational Linguistics ({ACL}-04)}, pages 478--485, Barcelona, Spain.

\bibitem[{Klein and Manning(2002)}]{klein-manning-2002-generative}
Dan Klein and Christopher~D. Manning. 2002.
\newblock \href {https://doi.org/10.3115/1073083.1073106} {A generative
  constituent-context model for improved grammar induction}.
\newblock In \emph{Proceedings of the 40th Annual Meeting of the Association
  for Computational Linguistics}, pages 128--135, Philadelphia, Pennsylvania,
  USA. Association for Computational Linguistics.

\bibitem[{Li et~al.(2020)Li, Kim, Amplayo, and Keller}]{li-etal-2020-heads}
Bowen Li, Taeuk Kim, Reinald~Kim Amplayo, and Frank Keller. 2020.
\newblock \href {https://aclanthology.org/2020.aacl-main.43} {Heads-up!
  unsupervised constituency parsing via self-attention heads}.
\newblock In \emph{Proceedings of the 1st Conference of the Asia-Pacific
  Chapter of the Association for Computational Linguistics and the 10th
  International Joint Conference on Natural Language Processing}, pages
  409--424, Suzhou, China. Association for Computational Linguistics.

\bibitem[{Liu et~al.(2023)Liu, Yang, Kim, and Tu}]{liu-etal-2023-simple}
Wei Liu, Songlin Yang, Yoon Kim, and Kewei Tu. 2023.
\newblock \href {https://doi.org/10.18653/v1/2023.findings-emnlp.113} {Simple
  hardware-efficient {PCFG}s with independent left and right productions}.
\newblock In \emph{Findings of the Association for Computational Linguistics:
  EMNLP 2023}, pages 1662--1669, Singapore. Association for Computational
  Linguistics.

\bibitem[{Liu et~al.(2019)Liu, Ott, Goyal, Du, Joshi, Chen, Levy, Lewis,
  Zettlemoyer, and Stoyanov}]{DBLP:journals/corr/abs-1907-11692}
Yinhan Liu, Myle Ott, Naman Goyal, Jingfei Du, Mandar Joshi, Danqi Chen, Omer
  Levy, Mike Lewis, Luke Zettlemoyer, and Veselin Stoyanov. 2019.
\newblock \href {https://arxiv.org/abs/1907.11692} {Roberta: {A} robustly
  optimized {BERT} pretraining approach}.
\newblock \emph{CoRR}, abs/1907.11692.

\bibitem[{Loshchilov and Hutter(2019)}]{loshchilov2018decoupled}
Ilya Loshchilov and Frank Hutter. 2019.
\newblock \href {https://openreview.net/forum?id=Bkg6RiCqY7} {Decoupled weight
  decay regularization}.
\newblock In \emph{International Conference on Learning Representations}.

\bibitem[{Marcus et~al.(1993)Marcus, Santorini, and
  Marcinkiewicz}]{marcus-etal-1993-building}
Mitchell~P. Marcus, Beatrice Santorini, and Mary~Ann Marcinkiewicz. 1993.
\newblock \href {https://aclanthology.org/J93-2004} {Building a large annotated
  corpus of {E}nglish: The {P}enn {T}reebank}.
\newblock \emph{Computational Linguistics}, 19(2):313--330.

\bibitem[{Mare{\v{c}}ek and Rosa(2018)}]{marecek-rosa-2018-extracting}
David Mare{\v{c}}ek and Rudolf Rosa. 2018.
\newblock \href {https://doi.org/10.18653/v1/W18-5444} {Extracting syntactic
  trees from transformer encoder self-attentions}.
\newblock In \emph{Proceedings of the 2018 {EMNLP} Workshop {B}lackbox{NLP}:
  Analyzing and Interpreting Neural Networks for {NLP}}, pages 347--349,
  Brussels, Belgium. Association for Computational Linguistics.

\bibitem[{Mare{\v{c}}ek and Rosa(2019)}]{marecek-rosa-2019-balustrades}
David Mare{\v{c}}ek and Rudolf Rosa. 2019.
\newblock \href {https://doi.org/10.18653/v1/W19-4827} {From balustrades to
  pierre vinken: Looking for syntax in transformer self-attentions}.
\newblock In \emph{Proceedings of the 2019 ACL Workshop BlackboxNLP: Analyzing
  and Interpreting Neural Networks for NLP}, pages 263--275, Florence, Italy.
  Association for Computational Linguistics.

\bibitem[{Maveli and Cohen(2022)}]{maveli-cohen-2022-co}
Nickil Maveli and Shay Cohen. 2022.
\newblock \href {https://doi.org/10.18653/v1/2022.findings-acl.101}
  {{C}o-training an {U}nsupervised {C}onstituency {P}arser with {W}eak
  {S}upervision}.
\newblock In \emph{Findings of the Association for Computational Linguistics:
  ACL 2022}, pages 1274--1291, Dublin, Ireland. Association for Computational
  Linguistics.

\bibitem[{Mikolov et~al.(2013{\natexlab{a}})Mikolov, Sutskever, Chen, Corrado,
  and Dean}]{NIPS2013_9aa42b31}
Tomas Mikolov, Ilya Sutskever, Kai Chen, Greg~S Corrado, and Jeff Dean.
  2013{\natexlab{a}}.
\newblock \href
  {https://proceedings.neurips.cc/paper_files/paper/2013/file/9aa42b31882ec039965f3c4923ce901b-Paper.pdf}
  {Distributed representations of words and phrases and their
  compositionality}.
\newblock In \emph{Advances in Neural Information Processing Systems},
  volume~26. Curran Associates, Inc.

\bibitem[{Mikolov et~al.(2013{\natexlab{b}})Mikolov, Yih, and
  Zweig}]{mikolov-etal-2013-linguistic}
Tomas Mikolov, Wen-tau Yih, and Geoffrey Zweig. 2013{\natexlab{b}}.
\newblock \href {https://aclanthology.org/N13-1090} {Linguistic regularities in
  continuous space word representations}.
\newblock In \emph{Proceedings of the 2013 Conference of the North {A}merican
  Chapter of the Association for Computational Linguistics: Human Language
  Technologies}, pages 746--751, Atlanta, Georgia. Association for
  Computational Linguistics.

\bibitem[{Paszke et~al.(2019)Paszke, Gross, Massa, Lerer, Bradbury, Chanan,
  Killeen, Lin, Gimelshein, Antiga, Desmaison, Kopf, Yang, DeVito, Raison,
  Tejani, Chilamkurthy, Steiner, Fang, Bai, and
  Chintala}]{NEURIPS2019_bdbca288}
Adam Paszke, Sam Gross, Francisco Massa, Adam Lerer, James Bradbury, Gregory
  Chanan, Trevor Killeen, Zeming Lin, Natalia Gimelshein, Luca Antiga, Alban
  Desmaison, Andreas Kopf, Edward Yang, Zachary DeVito, Martin Raison, Alykhan
  Tejani, Sasank Chilamkurthy, Benoit Steiner, Lu~Fang, Junjie Bai, and Soumith
  Chintala. 2019.
\newblock \href
  {https://proceedings.neurips.cc/paper_files/paper/2019/file/bdbca288fee7f92f2bfa9f7012727740-Paper.pdf}
  {Pytorch: An imperative style, high-performance deep learning library}.
\newblock In \emph{Advances in Neural Information Processing Systems},
  volume~32. Curran Associates, Inc.

\bibitem[{Radford(2018)}]{radford2018improving}
Alec Radford. 2018.
\newblock \href
  {https://cdn.openai.com/research-covers/language-unsupervised/language_understanding_paper.pdf}
  {Improving language understanding by generative pre-training}.
\newblock \emph{OpenAI}.

\bibitem[{Roark(2001)}]{roark-2001-probabilistic}
Brian Roark. 2001.
\newblock \href {https://doi.org/10.1162/089120101750300526} {Probabilistic
  top-down parsing and language modeling}.
\newblock \emph{Computational Linguistics}, 27(2):249--276.

\bibitem[{Shayegh et~al.(2024)Shayegh, Cao, Zhu, Cheung, and
  Mou}]{shayegh2024ensemble}
Behzad Shayegh, Yanshuai Cao, Xiaodan Zhu, Jackie~CK Cheung, and Lili Mou.
  2024.
\newblock \href {https://openreview.net/forum?id=RR8y0WKrFv} {Ensemble
  distillation for unsupervised constituency parsing}.
\newblock In \emph{The Twelfth International Conference on Learning
  Representations}.

\bibitem[{Shen et~al.(2018{\natexlab{a}})Shen, Lin, Jacob, Sordoni, Courville,
  and Bengio}]{shen-etal-2018-straight}
Yikang Shen, Zhouhan Lin, Athul~Paul Jacob, Alessandro Sordoni, Aaron
  Courville, and Yoshua Bengio. 2018{\natexlab{a}}.
\newblock \href {https://doi.org/10.18653/v1/P18-1108} {Straight to the tree:
  Constituency parsing with neural syntactic distance}.
\newblock In \emph{Proceedings of the 56th Annual Meeting of the Association
  for Computational Linguistics (Volume 1: Long Papers)}, pages 1171--1180,
  Melbourne, Australia. Association for Computational Linguistics.

\bibitem[{Shen et~al.(2018{\natexlab{b}})Shen, Lin, wei Huang, and
  Courville}]{shen2018neural}
Yikang Shen, Zhouhan Lin, Chin wei Huang, and Aaron Courville.
  2018{\natexlab{b}}.
\newblock \href {https://openreview.net/forum?id=rkgOLb-0W} {Neural language
  modeling by jointly learning syntax and lexicon}.
\newblock In \emph{International Conference on Learning Representations}.

\bibitem[{Shen et~al.(2019)Shen, Tan, Sordoni, and Courville}]{shen2018ordered}
Yikang Shen, Shawn Tan, Alessandro Sordoni, and Aaron Courville. 2019.
\newblock \href {https://openreview.net/forum?id=B1l6qiR5F7} {Ordered neurons:
  Integrating tree structures into recurrent neural networks}.
\newblock In \emph{International Conference on Learning Representations}.

\bibitem[{Shen et~al.(2021)Shen, Tay, Zheng, Bahri, Metzler, and
  Courville}]{shen-etal-2021-structformer}
Yikang Shen, Yi~Tay, Che Zheng, Dara Bahri, Donald Metzler, and Aaron
  Courville. 2021.
\newblock \href {https://doi.org/10.18653/v1/2021.acl-long.559}
  {{S}truct{F}ormer: Joint unsupervised induction of dependency and
  constituency structure from masked language modeling}.
\newblock In \emph{Proceedings of the 59th Annual Meeting of the Association
  for Computational Linguistics and the 11th International Joint Conference on
  Natural Language Processing (Volume 1: Long Papers)}, pages 7196--7209,
  Online. Association for Computational Linguistics.

\bibitem[{Stern et~al.(2017)Stern, Andreas, and
  Klein}]{stern-etal-2017-minimal}
Mitchell Stern, Jacob Andreas, and Dan Klein. 2017.
\newblock \href {https://doi.org/10.18653/v1/P17-1076} {A minimal span-based
  neural constituency parser}.
\newblock In \emph{Proceedings of the 55th Annual Meeting of the Association
  for Computational Linguistics (Volume 1: Long Papers)}, pages 818--827,
  Vancouver, Canada. Association for Computational Linguistics.

\bibitem[{Tillet et~al.(2019)Tillet, Kung, and Cox}]{10.1145/3315508.3329973}
Philippe Tillet, H.~T. Kung, and David Cox. 2019.
\newblock \href {https://doi.org/10.1145/3315508.3329973} {Triton: an
  intermediate language and compiler for tiled neural network computations}.
\newblock In \emph{Proceedings of the 3rd ACM SIGPLAN International Workshop on
  Machine Learning and Programming Languages}, MAPL 2019, page 10–19, New
  York, NY, USA. Association for Computing Machinery.

\bibitem[{Wang et~al.(2019)Wang, Lee, and Chen}]{wang-etal-2019-tree}
Yaushian Wang, Hung-Yi Lee, and Yun-Nung Chen. 2019.
\newblock \href {https://doi.org/10.18653/v1/D19-1098} {Tree transformer:
  Integrating tree structures into self-attention}.
\newblock In \emph{Proceedings of the 2019 Conference on Empirical Methods in
  Natural Language Processing and the 9th International Joint Conference on
  Natural Language Processing (EMNLP-IJCNLP)}, pages 1061--1070, Hong Kong,
  China. Association for Computational Linguistics.

\bibitem[{Wang and Utiyama(2024)}]{wang-utiyama-cont-disc}
Yiran Wang and Masao Utiyama. 2024.
\newblock \href {https://arxiv.org/abs/2406.07812} {To be continuous, or to be
  discrete, those are bits of questions}.
\newblock In \emph{Proceedings of The 62nd Annual Meeting of the Association
  for Computational Linguistics}, Bangkok, Thailand. Association for
  Computational Linguistics.

\bibitem[{Wang et~al.(2023)Wang, Watanabe, Utiyama, and
  Matsumoto}]{wang-etal-2023-24}
Yiran Wang, Taro Watanabe, Masao Utiyama, and Yuji Matsumoto. 2023.
\newblock \href {https://aclanthology.org/2023.ijcnlp-main.27} {24-bit
  languages}.
\newblock In \emph{Proceedings of the 13th International Joint Conference on
  Natural Language Processing and the 3rd Conference of the Asia-Pacific
  Chapter of the Association for Computational Linguistics (Volume 1: Long
  Papers)}, pages 408--419, Nusa Dua, Bali. Association for Computational
  Linguistics.

\bibitem[{Wolf et~al.(2020)Wolf, Debut, Sanh, Chaumond, Delangue, Moi, Cistac,
  Rault, Louf, Funtowicz, Davison, Shleifer, von Platen, Ma, Jernite, Plu, Xu,
  Le~Scao, Gugger, Drame, Lhoest, and Rush}]{wolf-etal-2020-transformers}
Thomas Wolf, Lysandre Debut, Victor Sanh, Julien Chaumond, Clement Delangue,
  Anthony Moi, Pierric Cistac, Tim Rault, Remi Louf, Morgan Funtowicz, Joe
  Davison, Sam Shleifer, Patrick von Platen, Clara Ma, Yacine Jernite, Julien
  Plu, Canwen Xu, Teven Le~Scao, Sylvain Gugger, Mariama Drame, Quentin Lhoest,
  and Alexander Rush. 2020.
\newblock \href {https://doi.org/10.18653/v1/2020.emnlp-demos.6} {Transformers:
  State-of-the-art natural language processing}.
\newblock In \emph{Proceedings of the 2020 Conference on Empirical Methods in
  Natural Language Processing: System Demonstrations}, pages 38--45, Online.
  Association for Computational Linguistics.

\bibitem[{Xue et~al.(2005)Xue, Xia, Chiou, and
  Palmer}]{XUE_XIA_CHIOU_PALMER_2005}
Naiwen Xue, Fei Xia, Fu-Dong Chiou, and Marta Palmer. 2005.
\newblock \href {https://doi.org/10.1017/S135132490400364X} {The penn chinese
  treebank: Phrase structure annotation of a large corpus}.
\newblock \emph{Natural Language Engineering}, 11(2):207–238.

\bibitem[{Yang et~al.(2022)Yang, Liu, and Tu}]{yang-etal-2022-dynamic}
Songlin Yang, Wei Liu, and Kewei Tu. 2022.
\newblock \href {https://doi.org/10.18653/v1/2022.naacl-main.353} {Dynamic
  programming in rank space: Scaling structured inference with low-rank {HMM}s
  and {PCFG}s}.
\newblock In \emph{Proceedings of the 2022 Conference of the North American
  Chapter of the Association for Computational Linguistics: Human Language
  Technologies}, pages 4797--4809, Seattle, United States. Association for
  Computational Linguistics.

\bibitem[{Yang et~al.(2021{\natexlab{a}})Yang, Zhao, and
  Tu}]{yang-etal-2021-neural}
Songlin Yang, Yanpeng Zhao, and Kewei Tu. 2021{\natexlab{a}}.
\newblock \href {https://doi.org/10.18653/v1/2021.acl-long.209} {Neural
  bi-lexicalized {PCFG} induction}.
\newblock In \emph{Proceedings of the 59th Annual Meeting of the Association
  for Computational Linguistics and the 11th International Joint Conference on
  Natural Language Processing (Volume 1: Long Papers)}, pages 2688--2699,
  Online. Association for Computational Linguistics.

\bibitem[{Yang et~al.(2021{\natexlab{b}})Yang, Zhao, and
  Tu}]{yang-etal-2021-pcfgs}
Songlin Yang, Yanpeng Zhao, and Kewei Tu. 2021{\natexlab{b}}.
\newblock \href {https://doi.org/10.18653/v1/2021.naacl-main.117} {{PCFG}s can
  do better: Inducing probabilistic context-free grammars with many symbols}.
\newblock In \emph{Proceedings of the 2021 Conference of the North American
  Chapter of the Association for Computational Linguistics: Human Language
  Technologies}, pages 1487--1498, Online. Association for Computational
  Linguistics.

\bibitem[{Yu et~al.(2020)Yu, Bohnet, and Poesio}]{yu-etal-2020-named}
Juntao Yu, Bernd Bohnet, and Massimo Poesio. 2020.
\newblock \href {https://doi.org/10.18653/v1/2020.acl-main.577} {Named entity
  recognition as dependency parsing}.
\newblock In \emph{Proceedings of the 58th Annual Meeting of the Association
  for Computational Linguistics}, pages 6470--6476, Online. Association for
  Computational Linguistics.

\bibitem[{Zhang et~al.(2020)Zhang, Zhou, and Li}]{ijcai2020p0560}
Yu~Zhang, Houquan Zhou, and Zhenghua Li. 2020.
\newblock \href {https://doi.org/10.24963/ijcai.2020/560} {Fast and accurate
  neural crf constituency parsing}.
\newblock In \emph{Proceedings of the Twenty-Ninth International Joint
  Conference on Artificial Intelligence, {IJCAI-20}}, pages 4046--4053.
  International Joint Conferences on Artificial Intelligence Organization.
\newblock Main track.

\bibitem[{Zhao and Titov(2021)}]{zhao-titov-2021-empirical}
Yanpeng Zhao and Ivan Titov. 2021.
\newblock \href {https://aclanthology.org/2021.adaptnlp-1.17} {An empirical
  study of compound {PCFG}s}.
\newblock In \emph{Proceedings of the Second Workshop on Domain Adaptation for
  NLP}, pages 166--171, Kyiv, Ukraine. Association for Computational
  Linguistics.

\bibitem[{Zhu et~al.(2020)Zhu, Bisk, and Neubig}]{zhu-etal-2020-return}
Hao Zhu, Yonatan Bisk, and Graham Neubig. 2020.
\newblock \href {https://aclanthology.org/2020.tacl-1.42} {The return of
  lexical dependencies: Neural lexicalized {PCFG}s}.
\newblock \emph{Transactions of the Association for Computational Linguistics},
  8:647--661.

\end{thebibliography}

\appendix

\section{Datasets Statistics}
\label{sec:stats}

\begin{table}[!h]
\centering
  \adjustbox{width=\columnwidth}{
  \begin{tabular}{cccccc}
  \toprule
  \textsc{Dataset} & \textsc{Train} & \textsc{Dev} & \textsc{Test} & \textsc{Word} & \textsc{Span} \\
  \midrule
  \textsc{Ptb} & \num{39832} & \num{1700} & \num{2416} & \num{44363} & \num{8865092} \\
  \textsc{Ctb} & \num{18104} & \num{352} & \num{348} & \num{36800} & \num{6510230} \\
  \bottomrule
  \end{tabular}}
  \caption{Datasets statistics. Columns \textsc{Train}, \textsc{Dev}, and \textsc{Test} show the number of sentences in each split, while Columns \textsc{Word} and \textsc{Span} indicate the number of words and spans, respectively.
  } 
  \label{tab:stats}
\end{table}

\section{More Examples}
\label{sec:example}

\begin{figure}[h]
\centering
\adjustbox{scale=1.0}{
\begin{tikzpicture}
\centering
\Tree [.\code{1E56}
        [.\code{DEDA} Lucas ]
        [.\code{E5A4}
          [.\code{8844} brought ]
          [.\code{6E37}
            [.\code{2FB7}
              [.\code{CA28} the ]
              [.\code{0E35} groceries ] ]
            [.\code{65F7} [.\code{60E4} for ] [.\code{0E07} him ] ] ] ] ]
\end{tikzpicture}
}
\caption{Derivation example.}
\end{figure}

\begin{figure}[h]
\centering
\adjustbox{scale=1.0}{
\begin{tikzpicture}
\centering
\Tree [.\code{4A04}
        [.\code{1CBF}
          [.\code{188A} She ]
          [.\code{E5A7}
            [.\code{8844} ate ]
            [.\code{2FB7}
              [.\code{CA28} the ]
              [.\code{0E27} pumpkin ] ] ] ]
        [.\code{098F}
          [.\code{488F} [.\code{408F} that ] [.\code{4B0F} Luna ] ]
          [.\code{8104} smashed ] ] ]
\end{tikzpicture}
}
\caption{Derivation example.}
\end{figure}

\begin{figure*}[h]
\centering
\adjustbox{scale=1.0}{
\begin{tikzpicture}
\centering
\Tree [.\code{4E76}
        [.\code{FFBF} [.\code{DABA} The ] [.\code{4E6E} council ] ]
        [.\code{E5B4}
          [.\code{8044} approved ]
          [.\code{6E37}
            [.\code{2FB7}
              [.\code{CA28} the ]
              [.\code{0E27} proposal ] ]
            [.\code{65FF} [.\code{60E5} on ] [.\code{6E47} Monday ] ] ] ] ]
\end{tikzpicture}
}
\caption{Derivation example.}
\end{figure*}

\begin{figure*}[h]
\centering
\adjustbox{scale=1.0}{
\begin{tikzpicture}
\centering
\Tree [.\code{5E3B}
        [.\code{3FB7} [.\code{5A9A} All ] [.\code{0F07} prices ] ]
        [.\code{E1AF}
          [.\code{800D} are ]
          [.\code{6527}
            [.\code{8125} as ]
            [.\code{E6FF}
              [.\code{E265} of ]
              [.\code{6E76}
                [.\code{4E7A} monday ]
                [.\code{2FEF}
                  [.\code{CACB} 's ]
                  [.\code{0B65} close ] ] ] ] ] ] ]
\end{tikzpicture}
}
\caption{Derivation example.}
\end{figure*}

\begin{figure*}[h]
\centering
\adjustbox{scale=1.0}{
\begin{tikzpicture}
\centering
\Tree [.\code{1CAB}
        [.\code{188A} That ]
        [.\code{98AF}
          [.\code{C88A} 'll ]
          [.\code{1427}
            [.\code{9437}
              [.\code{8860} save ]
              [.\code{1DB7} [.\code{0823} us ] [.\code{0525} time ] ] ]
            [.\code{00EF}
              [.\code{424D} and ]
              [.\code{B4B7}
                [.\code{90A0} get ]
                [.\code{35A7}
                  [.\code{0827} people ]
                  [.\code{0705} involved ] ] ] ] ] ] ]
\end{tikzpicture}
}
\caption{Derivation example.}
\end{figure*}

\begin{figure*}[t]
\centering
\adjustbox{scale=1.0}{
\begin{tikzpicture}
\centering
\Tree [.\code{1E12}
        [.\code{3EBF} [.\code{5ABA} A ] [.\code{0E36} decision ] ]
        [.\code{B1AF}
          [.\code{884E} is ]
          [.\code{B5A3}
            [.\code{8882} n't ]
            [.\code{F531}
              [.\code{9131} expected ]
              [.\code{64A7}
                [.\code{20A5} until ]
                [.\code{6673}
                  [.\code{36F7}
                    [.\code{12A0} some ]
                    [.\code{2645} time ] ]
                  [.\code{67F7}
                    [.\code{42C1} next ]
                    [.\code{6647} year ] ] ] ] ] ] ] ]
\end{tikzpicture}
}
\caption{Derivation example.}
\end{figure*}

\begin{figure*}[t]
\centering
\adjustbox{scale=1.0}{
\begin{tikzpicture}
\centering
\Tree [.\code{2E50}
        [.\code{7E30}
          [.\code{3FF7}
            [.\code{5A98} Interest ]
            [.\code{0F15} expense ] ]
          [.\code{E4FB}
            [.\code{E0CC} in ]
            [.\code{6EFB}
              [.\code{CA21} the ]
              [.\code{7E63}
                [.\code{5E42} 1988 ]
                [.\code{7F67}
                  [.\code{5E40} third ]
                  [.\code{2A64} quarter ] ] ] ] ] ]
        [.\code{A4E1}
          [.\code{C005} was ]
          [.\code{3477}
            [.\code{1450} 75.3 ]
            [.\code{2475} million ] ] ] ]
\end{tikzpicture}
}
\caption{Derivation example.}
\end{figure*}

\begin{figure*}[t]
\centering
\adjustbox{scale=1.0}{
\begin{tikzpicture}
\centering
\Tree [.\code{0E50}
        [.\code{FFFA}
          [.\code{DE58} Resolution ]
          [.\code{EB6E}
            [.\code{8A4A} Funding ]
            [.\code{AB4E} Corp. ] ] ]
        [.\code{24AB}
          [.\code{400C} to ]
          [.\code{F4B1}
            [.\code{8030} sell ]
            [.\code{7671}
              [.\code{3473}
                [.\code{1050} 4.5 ]
                [.\code{2451} billion ] ]
              [.\code{7776}
                [.\code{7650} 30-year ]
                [.\code{2364} bonds ] ] ] ] ] ]
\end{tikzpicture}
}
\caption{Derivation example.}
\end{figure*}

\begin{figure*}[t]
\centering
\adjustbox{scale=1.0}{
\begin{tikzpicture}
\Tree [.\code{0E10}
        [.\code{6E7F}
          [.\code{7EFB}
            [.\code{5AB8} The ]
            [.\code{7F77}
              [.\code{5E03} following ]
              [.\code{4A47} month ] ] ]
          [.\code{EFBF} [.\code{CA28} the ] [.\code{4E6E} company ] ] ]
        [.\code{81AF}
          [.\code{C84D} put ]
          [.\code{B527}
            [.\code{8801} itself ]
            [.\code{2525}
              [.\code{0125} up ]
              [.\code{25F7}
                [.\code{60A0} for ]
                [.\code{0707} sale ] ] ] ] ] ]
\end{tikzpicture}
}
\caption{Derivation example.}
\end{figure*}

\begin{figure*}[t]
\centering
\adjustbox{width=\textwidth}{
\begin{tikzpicture}
\Tree [.\code{0A50}
        [.\code{4E54} [.\code{EE58} Dreyfus ] [.\code{4B8E} alone ] ]
        [.\code{A1AF}
          [.\code{884C} has ]
          [.\code{2601}
            [.\code{F5B3}
              [.\code{8040} seen ]
              [.\code{6FB3}
                [.\code{4AA1} its ]
                [.\code{7F23}
                  [.\code{1E01} money ]
                  [.\code{6767}
                    [.\code{4241} market ]
                    [.\code{0347} funds ] ] ] ] ]
            [.\code{2121}
              [.\code{0005} grow ]
              [.\code{2451}
                [.\code{24E3}
                  [.\code{60E5} from ]
                  [.\code{7635}
                    [.\code{3477}
                      [.\code{1010} 1 ]
                      [.\code{2435} billion ] ]
                    [.\code{65F7}
                      [.\code{E0E4} in ]
                      [.\code{6647} 1975 ] ] ] ]
                [.\code{24EF}
                  [.\code{602C} to ]
                  [.\code{2421}
                    [.\code{0401} closes ]
                    [.\code{24E3}
                      [.\code{602C} to ]
                      [.\code{7677}
                        [.\code{3477}
                          [.\code{1050} 15 ]
                          [.\code{2455} billion ] ]
                        [.\code{6245} today ] ] ] ] ] ] ] ] ] ]
\end{tikzpicture}
}
\caption{Derivation example.}
\end{figure*}

\begin{figure*}[t]
\centering
\adjustbox{width=\textwidth}{
\begin{tikzpicture}
\Tree [.\code{0E50}
        [.\code{6E56}
          [.\code{6E7E}
            [.\code{FF7A}
              [.\code{F4FA}
                [.\code{5098} At ]
                [.\code{FEBB}
                  [.\code{CA28} the ]
                  [.\code{767A}
                    [.\code{3672}
                      [.\code{1650} 932 ]
                      [.\code{2651} million ] ]
                    [.\code{FE7A}
                      [.\code{D658} T. ]
                      [.\code{EA4A}
                        [.\code{EE5A}
                          [.\code{FE58} Rowe ]
                          [.\code{EA4A} Price ] ]
                        [.\code{EA4A} High ] ] ] ] ] ]
              [.\code{AB4A} Yield ] ]
            [.\code{4B4E} Fund ] ]
          [.\code{4B0E} investors ] ]
        [.\code{F521}
          [.\code{8001} yanked ]
          [.\code{64A1}
            [.\code{00A1} out ]
            [.\code{7671}
              [.\code{34F3}
                [.\code{40A0} about ]
                [.\code{3577}
                  [.\code{1050} 182 ]
                  [.\code{2655} million ] ] ]
              [.\code{64E3}
                [.\code{60E4} in ]
                [.\code{6EF3}
                  [.\code{CA29} the ]
                  [.\code{7E63}
                    [.\code{5E41} past ]
                    [.\code{7777}
                      [.\code{5640} two ]
                      [.\code{2A44} months ] ] ] ] ] ] ] ] ]
\end{tikzpicture}
}
\caption{Derivation example.}
\end{figure*}

\begin{figure*}[t]
\centering
\adjustbox{width=\textwidth}{
\begin{tikzpicture}
\Tree [.\code{0E02}
        [.\code{4E37}
          [.\code{4E16}
            [.\code{F4FE}
              [.\code{D098} In ]
              [.\code{6EBF}
                [.\code{CA28} the ]
                [.\code{4E26} stands ] ] ]
            [.\code{4F0F} people ] ]
          [.\code{E7A7}
            [.\code{C284} waved ]
            [.\code{67A7} [.\code{4203} ANC ] [.\code{4B06} flags ] ] ] ]
        [.\code{E1A7}
          [.\code{C004} wore ]
          [.\code{0643}
            [.\code{77A3}
              [.\code{5201} ANC ]
              [.\code{6242} T-shirts ] ]
            [.\code{4601}
              [.\code{A5A7}
                [.\code{8004} sang ]
                [.\code{37A7}
                  [.\code{0203} ANC ]
                  [.\code{0345} songs ] ] ]
              [.\code{21EF}
                [.\code{424F} and ]
                [.\code{E5B3}
                  [.\code{8000} chanted ]
                  [.\code{2727}
                    [.\code{0203} ANC ]
                    [.\code{0345} slogans ] ] ] ] ] ] ] ]
\end{tikzpicture}
}
\caption{Derivation example.}
\end{figure*}

\begin{figure*}[t]
\centering
\adjustbox{width=\textwidth}{
\begin{tikzpicture}
\Tree [.\code{0E10}
        [.\code{4E30}
          [.\code{7EFF} [.\code{DABA} The ] [.\code{4E36} convoy ] ]
          [.\code{E0E5}
            [.\code{C044} of ]
            [.\code{24B3}
              [.\code{4080} about ]
              [.\code{37F7}
                [.\code{1201} 100 ]
                [.\code{0345} vehicles ] ] ] ] ]
        [.\code{E0AF}
          [.\code{C84D} was ]
          [.\code{6E33}
            [.\code{2CB7} [.\code{CA28} the ] [.\code{0E37} first ] ]
            [.\code{64AF}
              [.\code{602C} to ]
              [.\code{F423}
                [.\code{8020} make ]
                [.\code{6521}
                  [.\code{0535} deliveries ]
                  [.\code{6671}
                    [.\code{64BF}
                      [.\code{602C} to ]
                      [.\code{6FFF}
                        [.\code{CA28} the ]
                        [.\code{4E66} capital ] ] ]
                    [.\code{64F3}
                      [.\code{E0EC} in ]
                      [.\code{66F3}
                        [.\code{42E1} about ]
                        [.\code{7777}
                          [.\code{1641} 10 ]
                          [.\code{2A44} days ] ] ] ] ] ] ] ] ] ] ]
\end{tikzpicture}
}
\caption{Derivation example.}
\end{figure*}

\end{document}